\documentclass[10pt,journal,compsoc]{IEEEtran}
\ifCLASSOPTIONcompsoc
  \usepackage[nocompress]{cite}
\else
  \usepackage{cite}
\fi
\ifCLASSINFOpdf
\else
\fi
\usepackage{orcidlink}
\usepackage{algorithm}
\newcommand\tab[1][1cm]{\hspace*{#1}}
\usepackage{amsmath}
\usepackage{adjustbox}
\usepackage{multirow}
\usepackage{amssymb}
\usepackage{anyfontsize}
\usepackage{blindtext}
\usepackage{multirow}
\usepackage{amssymb}
\usepackage{pifont}
\usepackage[labelformat=simple]{subcaption}

\DeclareCaptionLabelFormat{subcaptionlabel}{\normalfont(\textbf{#2}\normalfont)}
\definecolor{myblue}{RGB}{0, 100, 200}
\usepackage{hyperref}

\usepackage{algpseudocode} 
\usepackage{booktabs} 
\usepackage{color, colortbl}
\usepackage{makecell}  

\definecolor{Gray}{gray}{0.9}
\definecolor{LightCyan}{rgb}{0.88,1,1}
\begin{document}
%
\title{Mitigating Degree Bias in Graph Representation Learning with Learnable Structural Augmentation and Structural Self-Attention}

%
%
%

\author{Van Thuy Hoang \orcidlink{0000-0003-4094-1123},
        Hyeon-Ju Jeon \orcidlink{0000-0002-2400-8360},
        and~O-Joun Lee \orcidlink{0000-0001-8921-5443}
\IEEEcompsocitemizethanks{\IEEEcompsocthanksitem Van Thuy Hoang and O-Joun Lee are with the Department of Artificial Intelligence, The Catholic University of Korea, 43, Jibong-ro, Bucheon-si, Gyeonggi-do 14662, Republic of Korea (email: hoangvanthuy90@catholic.ac.kr, ojlee@catholic.ac.kr)   
\IEEEcompsocthanksitem Hyeon-Ju Jeon is with Data Assimilation Group, Korea Institute of Atmospheric Prediction Systems (KIAPS), 35, Boramae-ro 5-gil, Dongjak-gu, Seoul 07071, Republic of Korea (email: hjjeon@kiaps.org)}}

\IEEEtitleabstractindextext{%
\begin{abstract}
Graph Neural Networks (GNNs) update node representations through message passing, which is primarily based on the homophily principle, assuming that adjacent nodes share similar features.
However, in real-world graphs with long-tailed degree distributions, high-degree nodes dominate message passing, causing a degree bias where low-degree nodes remain under-represented due to inadequate messages.
The main challenge in addressing degree bias is how to discover non-adjacent nodes to provide additional messages to low-degree nodes while reducing excessive messages for high-degree nodes. 
Nevertheless, exploiting non-adjacent nodes to provide valuable messages is challenging, as it could generate noisy information and disrupt the original graph structures.  
To solve it, we propose a novel Degree Fairness Graph Transformer, named DegFairGT, to mitigate degree bias by discovering structural similarities between non-adjacent nodes through learnable structural augmentation and structural self-attention.
Our key idea is to exploit non-adjacent nodes with similar roles in the same community to generate informative edges under our augmentation, which could provide informative messages between nodes with similar roles while ensuring that the homophily principle is maintained within the community.
By considering the structural similarities among non-adjacent nodes to generate informative edges, DegFairGT can overcome the imbalanced messages while still preserving the graph structures. 
To enable DegFairGT to learn such structural similarities, we then propose a structural self-attention to capture the similarities between node pairs. 
To preserve global graph structures and prevent graph augmentation from hindering graph structure, we propose a Self-Supervised Learning task to preserve p-step transition probability and regularize graph augmentation. 
Extensive experiments on six datasets showed that DegFairGT outperformed state-of-the-art baselines in degree fairness analysis, node classification, and node clustering tasks.

\end{abstract}


}

\maketitle

\IEEEdisplaynontitleabstractindextext

%
\IEEEpeerreviewmaketitle

\IEEEraisesectionheading{\section{Introduction}\label{sec:intro}}

\begin{figure}[t]
\centering
\includegraphics[width= .95\linewidth]{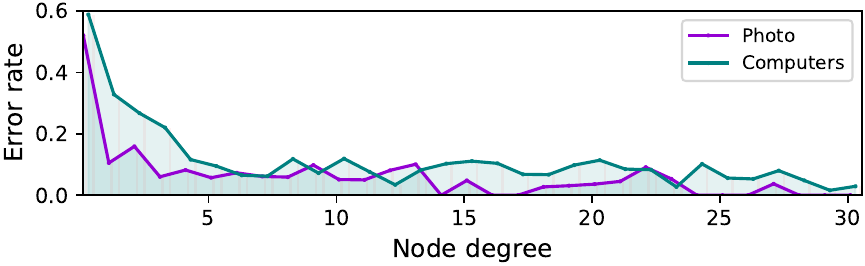}
\caption{Low-degree nodes are more misclassified than other nodes in the GT model \cite{dwivedi2020generalization} on Photo and Computers datasets.
The miss-classification rate is higher for low-degree nodes compared to high-degree nodes.
}
\label{fig:problem_b}
\end{figure}

Graphs provide a robust framework for capturing complex data by representing the relationships among entities in a wide range of domains, such as molecular structures and social networks \cite{li2024guest}.
However, real-world graphs often have long-tailed degree distributions, which could negatively affect GNNs' ability to learn low and high-degree nodes \cite{d098f7454ebe43e082f5e94c7ff051d3}.
That is, GNNs update node representations by aggregating information from neighboring nodes through message passing, following the homophily principle, which assumes that connected nodes tend to have similar features.
As a result, low-degree nodes receive a few messages from neighborhoods, which can cause bias or under-representation.
Meanwhile, high-degree nodes receive overabundant messages, leading to inherent limitations of message passing, e.g., over-smoothing problems \cite{HUANG2023110556}.
To illustrate this problem, Figure \ref{fig:problem_b} shows that GT~\cite{dwivedi2020generalization}, one of the recent graph transformer models, exhibits degree bias on Photo and Computers datasets \cite{DBLP:conf/sigir/McAuleyTSH15}. 
We can intuitively infer that the degree bias problem arises from insufficient messages from correlated nodes, not merely from the number of neighbors. 
This indicates that the graph topology could significantly influence the balance of messages under the message propagation in GNNs \cite{ji2023signal}.

To address this problem, most recent studies aim to deliver additional messages to low-degree nodes or remove redundant messages from high-degree nodes, mainly classified into three main groups: graph augmentation, observation range expansion, and neighborhood aggregation modulation.
The graph augmentation-based methods generate or remove edges randomly or heuristically to observe graphs with different views \cite{DBLP:conf/nips/Wang0SS22,DBLP:conf/cikm/LeeH0P22}.
The main idea is that augmentation schemes, e.g., edge perturbation, will continuously change the set of neighbors of each node and could help GNNs aggregate more messages to update node representations.
The graph augmentations involve modifying graph structures, e.g., adding or removing edges, to enhance model performance and address the degree bias problem. 
However, in real-world graphs, graph structures often remain stable, reflecting intrinsic relationships and dependencies among nodes \cite{artime2024robustness,ji2024focus}.

Some observation range expansion methods aim to expand the range of neighborhoods surrounding target nodes to capture high-order relationships and more information at each node representation update \cite{DBLP:conf/icml/Abu-El-HaijaPKA19,DBLP:conf/nips/YingCLZKHSL21,s23084168}.
Several studies aim to adjust the model weights to distill sufficient messages during the message-passing aggregation \cite{DBLP:conf/aaai/LiuN023}.
Nevertheless, as the degree bias originates from graph structure imbalance, aggregation modulation can not adapt properly to the varying size and structure graphs.

Although graph neural network strategies have shown effectiveness in mitigating degree bias, there are two limitations in existing studies that still limit their ability.
First, it is challenging to discover correlated nodes with high structural similarity to provide informative messages for each target node.
Most graph augmentation-based methods ignore the structural similarity between node pairs when forming edges, forcing GNNs to learn noisy and redundant messages.
The augmentation methods randomly add or drop edges to generate multiple graph views, which could mainly benefit low-degree nodes to obtain more messages from distant nodes \cite{DBLP:conf/cikm/LeeH0P22,DBLP:conf/www/0001XYLWW21,scgib}.
While randomly adding edges can bring more messages from non-adjacent nodes, the nodes could contain noisy and irrelevant information, which could force GNNs to aggregate redundant messages, resulting in suboptimal performance.
That is, pairs of nodes formed into edges should be strongly correlated to their structural similarity, which then could naturally enable GNNs to learn their representations and map them closely in the latent space.
Moreover, randomly dropping edges could severely degrade the low-degree node representations as they can be more under-represented due to the lack of messages.
\textbf{Second}, it is challenging to achieve both degree fairness and graph structure preservation simultaneously.
For augmentation-based methods, randomly adding edges can hinder preserving the original graph structures since any edge connecting two non-adjacent nodes forces GNNs to map node representations close significantly.
Besides, although several methods consider structural features \cite{DBLP:conf/www/0001XYLWW21}, their indistinction between generated and original edges causes loss of local connectivity and high-order proximity \cite{DBLP:conf/www/0001XYLWW21,DBLP:conf/nips/Wang0SS22,DBLP:conf/cikm/LeeH0P22}. 
Similarly, some observation range expansion methods ignore the distinctive messages between $k$-hop distances rooted at each node to its neighbors and even treat $k$-subgraph neighborhoods as complete graphs \cite{DBLP:conf/icml/ChenOB22,DBLP:conf/icml/Abu-El-HaijaPKA19}.
The loss of learning structural distance information and proximity between non-adjacent nodes hinders GNNs from preserving global graph structures.

In this study, we propose a novel framework, DegFairGT, to mitigate degree biases via structural graph augmentations and graph transformers.
The main idea is to exploit non-adjacent nodes with high structural similarity to form informative edges rather than random edge perturbation through graph augmentation, which could generate arbitrary edges and provide noisy information as well as disrupt the original graph structure.
More precisely, we first compute a structural similarity score for each node pair, which reflects how similar the two nodes are in terms of their graph structure.
The structural similarities are then used as a probability weight for sampling new edges under our augmentation.
By doing so, our augmentation can connect correlated non-adjacent nodes with high similarity, enabling more balanced messages and enabling the model to maintain the homophily principle. 
To enable DegFairGT to learn such structural similarities, we then propose a structural self-attention to capture the similarities between node pairs.
To preserve global graph structures and prevent graph augmentation from hindering graph structure, we propose a Self-Supervised Learning task to preserve p-step transition probability and regularize graph augmentation. 

\begin{figure}[t]
\centering
\includegraphics[width= .99\linewidth]{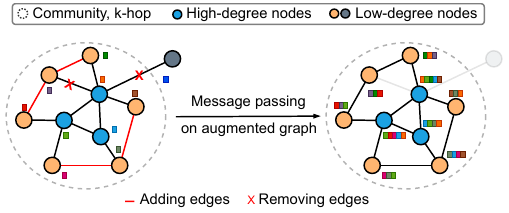}
\caption{
Structural graph augmentation adds intra-community edges between low-degree nodes and removes edges between nodes with different degrees or communities.
Given an input graph (left), our structural graph augmentation adds intra-community edges between low-degree nodes and removes edges between nodes with different degrees or communities.
Our structural augmentation can enable each node to obtain more valuable messages from neighbors within the community and k-hop distance through message passing (right).}
\label{fig:problem_a}
\end{figure}

The key challenge of designing structural graph augmentation is how to sample non-adjacent nodes to generate informative edges to mitigate degree biases while preserving intrinsic graph structures.
That is, when non-adjacent node pairs are connected through augmentation, it encourages GNNs to learn representations that map them closer in the latent space \cite{DBLP:conf/ijcai/ZhuoT22}. 
Hence, we introduce a sampling method to sample non-adjacent nodes with high similarities so that nodes with highly similar roles in the same community are sampled more frequently than other nodes.
These two aspects align with fundamental principles in graph structures, i.e., structural equivalence and the homophily principle, ensuring that the augmented graph could provide similar features for each pair of nodes while retaining global graph connectivity within a community \cite{DBLP:conf/iclr/PeiWCLY20}.
Intuitively, given a target node, we define the context nodes in order to be likely sampled to generate informative edges under structural augmentation \textit{w.r.t} (i) they are in the same clusters and within a $k $-hop distance, and (ii) they have a similar low-degree together.
The structural graph augmentation is end-to-end trainable through edge perturbation, making itself learn structural similarity between node pairs.
To illustrate our structural augmentation, Figure \ref{fig:problem_a} shows that it could generate more informative intra-community, mainly for low-degree nodes, and drop inter-community.

Although the augmentation module could generate informative edges, graph transformers are agnostic to learning the structural distance and proximity between non-adjacent nodes.
Note that nodes within $k$-hop distance within the same community tend to share dense connections, showing the high-order proximity that could be learned via self-attention.
Thus, we propose a structural self-attention to capture the high-order proximity and the node roles, leading to the model's ability to learn global graph structures.
The key idea is that we directly encode the high-order proximity into full dot product attention, enabling DegFairGT to discover the proximity and structure information while forming query and key matrices.

Our approach aims to address the degree bias problem while preserving the global graph topology without requiring any label information \cite{bai2024haqjsk}.
We achieve this by introducing a self-supervised learning task that preserves the global graph structure, utilizing the transition probability matrix for global graph structure preservation.
In addition, we note that the graph augmentation could collapse into insignificant edges and generate inappropriate graphs under the edge perturbation, resulting in graph transformers failing to learn representations.
For example, the augmentation module could learn to generate a fully connected graph or remove too many edges connecting adjacent nodes that retain no original graph structure. 
Such augmentations are not informative as they lose all the structural information from the original graph.
To mitigate this issue, we incorporate augmentation regularization as an additional self-supervised signal to ensure that the graph augmentation model still remains the original graph topology.
Specifically, we employ a binary cross-entropy (BCE) loss to minimize the dissimilarity between the augmented adjacency matrix and the original adjacency matrix, preventing the loss of the graph structure.
Finally, we obtain the embeddings extracted after the pre-training task for downstream tasks such as degree fairness analysis and node classification.
To the extent of our knowledge, DegFairGT is the first graph transformer model addressing degree biases based on graph augmentation.

Our contributions can be summarized as follows:
\begin{itemize}
\item We propose a structural graph augmentation, generating informative edges connecting non-adjacent nodes by exploring their correlation with high structural similarity. 
The generating edges between distant nodes could then assist the model in mitigating degree bias, as low-degree nodes can receive more messages from correlated nodes while messages for high-degree nodes will be reduced.

\item We propose structural self-attention to learn the relations between correlated nodes based on their structural similarity. 
The structural self-attention could capture the high-order proximity and the node roles, leading to the model’s ability to learn global graph structures.

\item We propose the SSL task to preserve graph structure and regularize the graph augmentation.
The SSL task can allow our model to reconstruct the global graph structure without using label information while using the augmentation module to mitigate degree bias.

\end{itemize}

\section{Related Work}

As degree bias originates from imbalanced distributions of the number of neighbors, several studies have proposed augmentation schemes to modify the neighborhoods \cite{DBLP:conf/nips/YouCSCWS20,DBLP:conf/nips/Wang0SS22,DBLP:journals/joi/LeeJJ21,nguyen2023companion}.
Early augmentation methods, e.g., GRACE~\cite{Zhu:2020vf} and RGRL~\cite{DBLP:conf/cikm/LeeH0P22}, randomly modify input graphs and then employ GNNs (Graph Neural Networks) with contrastive objectives to learn representations, not considering both proximity and structural similarity. 
GCA~\cite{DBLP:conf/www/0001XYLWW21} improves GRACE by using node centrality-based augmentation that remains edges connecting low-degree nodes and connecting high-degree nodes and randomly perturbs other edges, regardless of node proximity. 
GRADE~\cite{DBLP:conf/nips/Wang0SS22} samples $k$-subgraphs rooted in target nodes to generate edges connecting low-degree nodes within the $k$-subgraphs and removes inter-community edges based on node feature similarity.
Most heuristic augmentation methods randomly perturb the remaining edges not handled by their heuristics, despite the possibility of generating noisy edges.  
In contrast, we sample node pairs with high proximity and structural similarity using our learnable augmentation that can adapt to various graph structures.


The other approaches explore information in multi-hop neighborhoods surrounding each target node where node homophily is maintained \cite{Lee2020a,DBLP:journals/sensors/JeonCL22}.
One of the straightforward strategies is to stack more GNN layers to receive messages from a wider range of nodes \cite{DBLP:conf/iclr/XuHLJ19,velivckovic2017graph}.
Several studies expand sampling ranges for context nodes to capture more messages within $k$-hops at each layer \cite{DBLP:conf/nips/YingCLZKHSL21}.
SAT~\cite{DBLP:conf/icml/ChenOB22} uses $k$-subgraphs rooted in target nodes as contexts and employs a GNN layer to aggregate node features within the $k$-subgraphs.
Several well-known graph transformers, e.g., GT~\cite{dwivedi2020generalization} and SAN~\cite{DBLP:conf/nips/KreuzerBHLT21}, treat original graphs as fully-connected graphs to generate more messages from neighbors.
These models do not concern noisy signals from structurally dissimilar nodes nor consider connectivity among context nodes, leading to loss of high-order proximity. 
Graphformer \cite{DBLP:conf/nips/YingCLZKHSL21} considers $k$-hop distance between nodes in $k$-subgraphs by modifying self-attention.   
ANS\_GT \cite{DBLP:conf/nips/Zhang0HL22} employs multiple sampling strategies based on personalized PageRank, $k$-hop neighborhoods, and node feature similarity.
PSGT \cite{zhu2024propagation} extracts important edges between nodes under the information bottleneck principle. 
The model then could effectively capture multi-scale structures and long-sequence dependencies by utilizing a relational propagation graph as a novel position encoding. 
UGT \cite{DBLP:journals/corr/abs-2308-09517} generates edges connecting distant nodes with structural similarity and uses $k$-subgraph sampling together, regardless of community membership.
NodeFormer \cite{DBLP:conf/nips/WuZLWY22} undersamples one-hop neighborhoods learnably, which can help reduce noisy signals. 
EduCross \cite{li2024educross} integrates bipartite hypergraph learning with adversarial learning, allowing for effectively capturing high-order relationships between nodes.
HAQJSK \cite{bai2024haqjsk} aims to preserve hierarchical relationships in graphs by aligning structural features at multiple levels. This hierarchical approach improves representation power by considering local-to-global structural patterns.
over time, presenting challenges in capturing temporal dependencies and structural changes. The study introduces a novel method that leverages GCNs to effectively model these evolving graph structures.
DGCN \cite{gao2022novel} introduces a novel Graph Convolutional Network to effectively capture the global graph structure while maintaining local structural information by maximizing the agreement between global and local graph structure information. 
By leveraging node degrees, the model then could recognize the importance of neighboring to guide the message-passing aggregation.
In contrast, we discover correlated nodes that satisfy structural similarity to avoid noisy messages regarding homophily and structural equivalence.

Several studies have investigated the influence of degrees on the performance of GNNs to modulate effective message aggregation.
For example, DegFairGNN \cite{DBLP:conf/aaai/LiuN023} aims to modulate the neighborhood aggregation for each target node at each message passing layer to generate the debiasing context.
The idea is to distill information on the high-degree nodes and complement information on the low-degree node based on the node-degree information.
Kojaku et. al. \cite{DBLP:conf/nips/KojakuYCA21} focuses on counteracting the sampling of nodes in random walks.
That is, low-degree nodes will be sampled more based on the random walks strategy, which can benefit the low-degree node representations.
Liu et. al. \cite{DBLP:conf/kdd/LiuN021} aims to robust low-degree node representations by translating neighborhood information from high-degree to low-degree adjacent nodes.
The idea of GAUG \cite{DBLP:conf/aaai/0003LNW0S21} is to preserve the intrinsic graph structure while perturbing edges connecting low-degree nodes.
Jin et. al. \cite{DBLP:conf/kdd/Jin0LTWT20} addresses the low-degree bias directly on perturbed graphs to learn robust representations while resisting adversarial influence.
RawlsGCN \cite{DBLP:conf/www/KangZXLT22} focuses on exploring the relations between the node degree and the gradient of the weighted matrix at each message passing layer to mitigate the model performance disparity.
However, most existing studies focus on the structural difference between nodes to improve the model performance rather than address the origin of the degree bias.






\begin{figure*}[t] 
\centering 
 \includegraphics[width=.99 \linewidth]{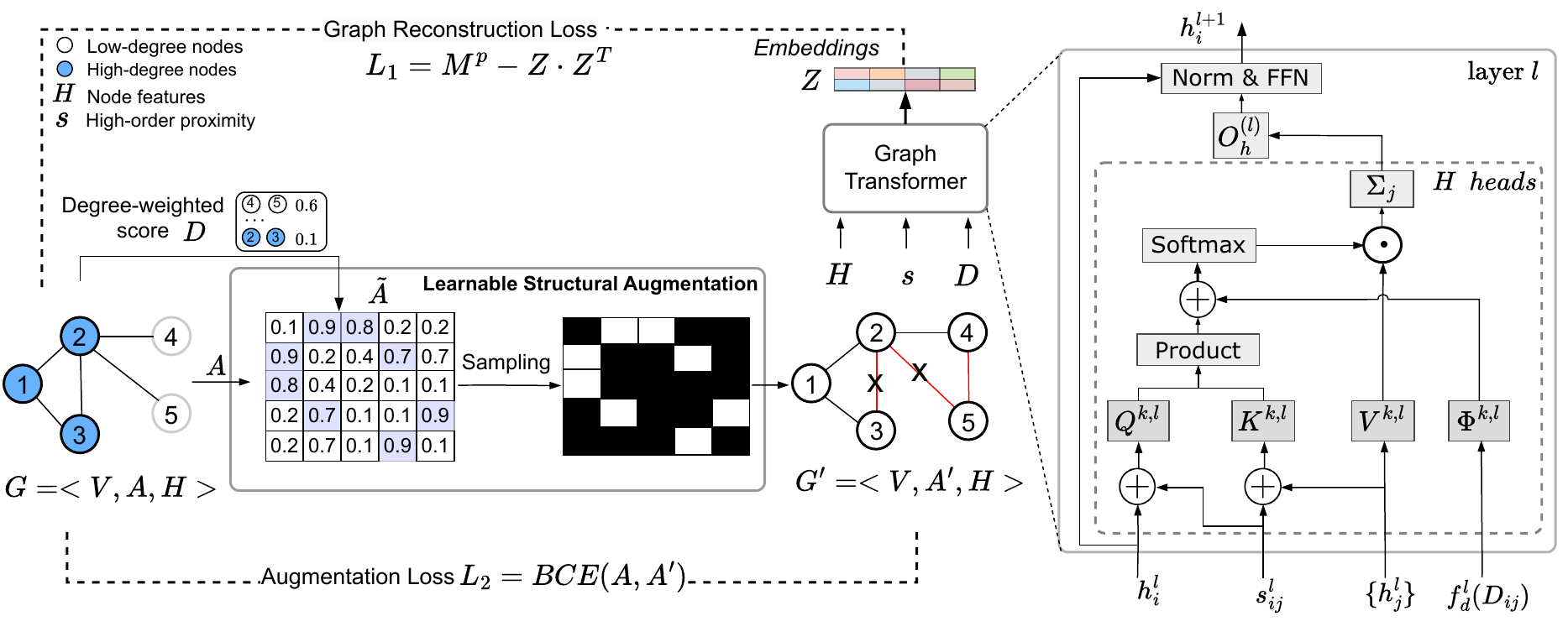}
 \caption{The overall architecture of DegFairGT.
DegFairGT comprises two main blocks: structural graph augmentation and structural self-attention. 
The augmentation module takes the original adjacency matrix $A$ and the degree-weighted score $D$ as inputs and then samples edges with probabilities defined in $\tilde{A}$ to generate a new graph $G'$.
After the augmentation, the new graph $G'$ is fed into the graph transformers networks, which then learn the representations $Z$.
The self-attention module at layer $l$-th receives a node feature of the target node $h^{l}_i$, the features of neighbouring nodes $\{h^{l}_j\}$, the proximity $s^{l}_ij$, and the structural similarity $f^{l}_d(D_{ij})$ between nodes $v_i$ and $v_j$ as inputs.
Finally, the learned representation $Z$ could be used for various downstream tasks, e.g., node classification.}
 \label{fig:model}
\end{figure*}

\section{Methodology}

This section first introduces our learnable structural graph augmentation (Sect. \ref{subsec:AutomatedGraphAugmentation}), explains the DegFairGT architecture in detail (Sect. \ref{subsec:graphtransformerarchitecture}), and
finally, introduces the SSL task for graph structure preservation (Sect. \ref{subsec:ssl}). 
Figure \ref{fig:model} shows the overall architecture of our framework.

\subsection{Learnable Structural Graph Augmentation}
\label{subsec:AutomatedGraphAugmentation}

Our structural augmentation aims to explore correlated nodes that can provide informative messages to learn degree-unbiased representations. 
Let $G =\langle V, A, H \rangle$ denote an input graph, 
where $V$ refers to the set of nodes, 
$A$ indicates the adjacency matrix,
and $H$ denotes the initial node features.
We first compose a modified adjacency matrix $\tilde{A}$ by a linear combination of a degree-weighted matrix $D$ and the original adjacency matrix $A$. 
The matrix $D$ quantifies the structural similarity between pairs of nodes without degree bias.
Then, we obtain an augmented graph $G' =\langle V, A', H\rangle$ by applying edge perturbation to $\tilde{A}$ as: 
\begin{equation}
\label{eq:d}
A' = \mathcal{T}_A (\tilde{A}), \tab  \tilde{A}=\xi A +\zeta   D\ ,
\end{equation} 
where $\xi$ and $\zeta$ are hyper-parameters, and $\mathcal{T}_A(\cdot)$ indicates the edge perturbation that is learnable.
We now explain the strategies for constructing $D$ and $\mathcal{T}_A(\cdot)$.



\subsubsection{Capturing Global Context Nodes in Community}
\label{subsubsection:Capturingglobalcontextnodes}

\begin{figure}
\centering
\includegraphics[width= .96\linewidth]{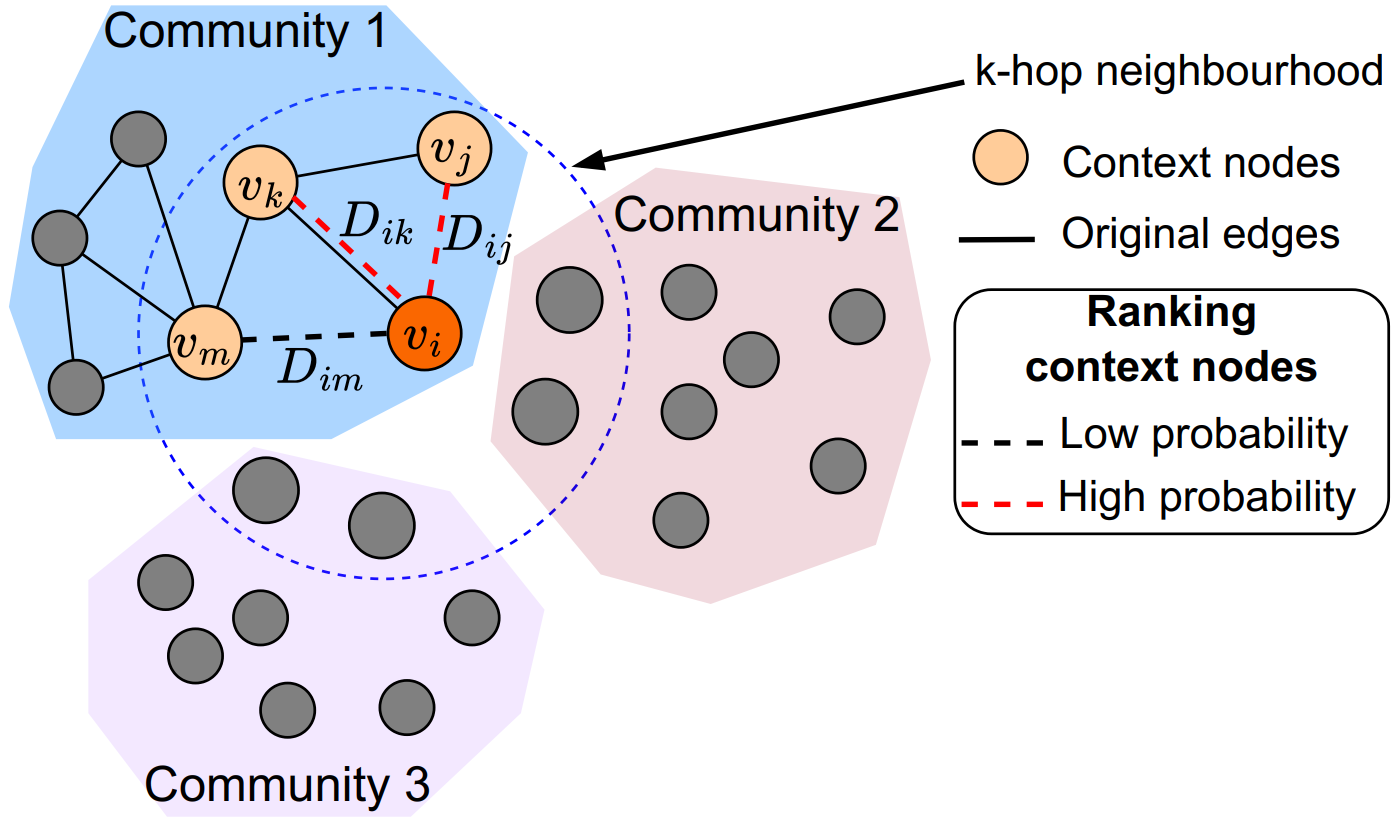}
\caption{
For a target node $v_i$, our augmentation samples context nodes in the same community and ranks the context nodes based on their node degrees within $k$-hops.
Two nodes  $v_k $ and  $v_j $ have a high correlation with the target node $v_i$ as they have similar low degrees, respectively.
}
\label{fig:sampling}
\end{figure}


Note that we aim to solve the degree bias problem by delivering messages between nodes with high structural similarity. 
In graphs, nodes in the same community tend to share globally similar features, showing the homophily principle that could benefit models from exploring structural similarity between non-adjacent nodes.
That is, we aim to explore the set of non-adjacent node candidates within the same community, which then will be likely to be sampled to generate informative edges via our augmentation.
We argue that homophily and structural equivalence between non-adjacent nodes can be explored to generate informative edges connecting correlated nodes in terms of their proximity and structural similarity \cite{DBLP:conf/iclr/PeiWCLY20}. 
Intuitively, we aim to explore context nodes that share the global semantic information, i.e., similar features, with the target nodes through the clustering.
We then refine the context nodes by removing nodes that are not reachable from the target nodes within $k$-hops as far-apart nodes do not have high correlation and proximity even within the same community. 
Specifically, we cluster nodes in a graph into $M$ communities ($G =\{G_1, G_2, \cdots, G_M\}$) by applying the $k$-means clustering to node initial features \cite{DBLP:conf/aaai/Lee0P22,DBLP:conf/icml/00080XZYLL23}. 
We then define the context nodes of a target node $v_i \in G_m$ as:
\begin{eqnarray}
\label{neighbour_set}
N(v_i) =\left\{ v_j\in V \Big\lvert A_{ij}^{(k)}> 0\ , v_j \in G_m\right\},
\end{eqnarray}
where $A^{(k)} = \sum_{l=1}^k A^l$ is the transition matrix of $G$ at $k$-step. 
$A_{ij}^{(k)} > 0$ denotes the reachability between $v_i$ and $v_j$ within $k$-hops.

\subsubsection{Ranking Context Nodes}
\label{subsubsection:Capturingsimilaritywithlowdegreerelatedbiases}

We note that the set of context nodes can be large as nodes are mostly connected densely in the community, even within $k$-hop distance.
This could make it challenging to discover nodes with high structural similarity to generate informative edges for each target node.
To solve it, we rank the context nodes based on their structural similarity to the target nodes with a low-degree bias.
The key idea is that low-degree nodes are likely to be sampled more to generate edges than high-degree nodes that already have many edges.
Thus, we assign higher priorities to node pairs with similar degrees than with different degrees and low- than high-degree nodes. 
Specifically, given a target node $v_i$, we rank the context nodes $N(v_i) \ni v_j$ based on their priorities quantified by the inverse of their node degrees, ${1}/{d_i}$ and ${1}/{d_j}$, as: 
\begin{equation}
\label{degree_bias}
\resizebox{0.43 \textwidth}{!}{$ D_{ij}=\frac{1}{\sqrt{d_i\cdot d_j}}, \text{when } v_j\in  N(v_i), D_{ij}=0, \text{otherwise}, $}
\end{equation}
where $d_i$ and  $d_j$ denote the degree of nodes $v_i$ and $v_j$, respectively, and $D_{ij}$ indicates degree-weighted score between $v_i$ and $v_j$. 
By doing so, edges connecting nodes with similar low degrees could be sampled with a high probability.
Figure \ref{fig:sampling} illustrates our augmentation strategy with the target node $v_i$ and three communities.
We first sample $v_j$, $v_k$, and $v_m$ as context nodes of $v_i$ since they are in the same community with $v_i$ and within $k$-hop distance from $v_i$.
Then, we use Equation \ref{degree_bias} to rank the context nodes based on their similarity in terms of roles.
We then linearly combine the original adjacency matrix ($A$) with the degree-weighted score matrix ($D$) consisting of $D_{ij}$ for every node pair to generate $\tilde{A}$.
In a nutshell, given a target node  $v_i$, we sample a set of context nodes $N(v_i)$ from Equation \ref{neighbour_set}, and then compute the degree-weighted score towards low-degree nodes based on Equation \ref{degree_bias}.

\begin{figure}[tb]
\centering 
\includegraphics[width= 0.8 \linewidth]{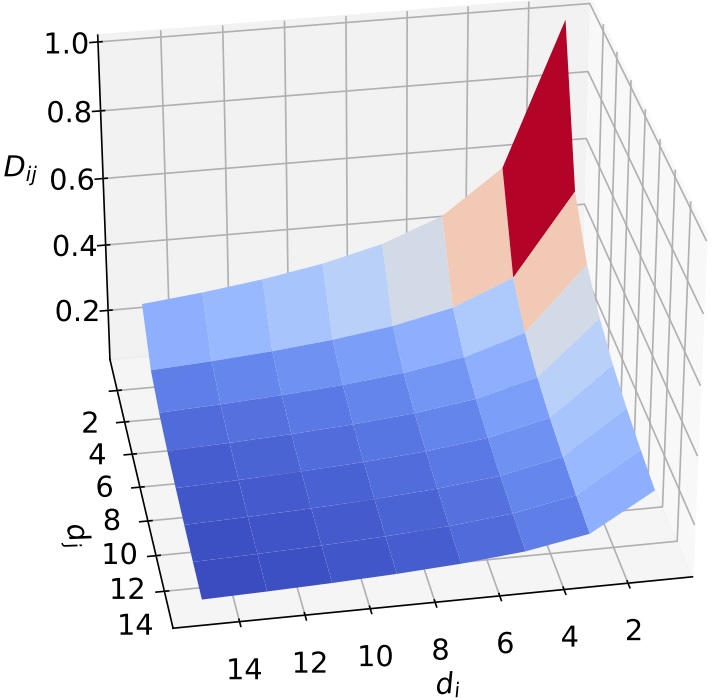}
\caption{An analysis on degree-weighted matrix $D$.
When two nodes $v_i$ and $v_j$ have low degrees together, they are more frequently sampled to generate valuable edges.}
\label{fig:dij}
\end{figure}

To illustrate the sampling bias of the degree-weighted matrix $D$, we conduct a visualization study, as shown in Figure \ref{fig:dij}.
This analysis delivers how different node degrees influence edge sampling in our model.
We observed that when $v_i$ and $v_j$ are low-degree together, $D_{ij}$ will be large and close to 1, which can be sampled more frequently to make valuable edges for low-degree nodes.
This indicates that low-degree nodes will be sampled more frequently, thereby alleviating the degree problem.
In contrast, when two nodes have high-degree together, the values $D_{ij}$ will be close to 0, which is less sampled to make edges.
When a node $v_i$ has a high degree and node $v_j$ has a low degree, the value $D_{ij}$ is still relatively small, leading to less frequent sampling compared to low-degree pairs.




\subsubsection{Edge Perturbation}
\label{subsubsection:Edgeperturbation}

To generate augmented graphs, we then transform $\tilde{A}$ into a new adjacency matrix $A'$ by sampling with the Bernoulli distribution parameterized with the probability in $\tilde{A}$, as:
\begin{equation}
\label{Bernoulli}
   A_{ij}^{'}=\text{Bernoulli}\left(\tilde{A}_{ij}\right) .
\end{equation}
To make the Bernoulli sampling function differentiable in an end-to-end manner, we utilize the commonly used scheme to approximate the Bernoulli sampling in Equation \ref{Bernoulli}.
Specifically, we approximate the Bernoulli sampling process by using Gumbel-Softmax as a re-parameterization trick to relax the discrete distribution \cite{jang2017categorical,maddison2017the}.

\subsection{Graph Transformer Architecture}
\label{subsec:graphtransformerarchitecture}

Given an input graph $G$, the node feature $x_{i}\in R^{d_{0}\times 1}$ of node $v_{i}$ is first projected via a linear transformation to $d$-dimensional hidden vector $h_{i}^{0}$, as:
\begin{eqnarray}
    h_{i}^{0}={{W}_{0}}{{x}_{i}}+{{b}_{0}},
\end{eqnarray}
where ${W}_{0} \in R^{d\times d_{0}} $ and  ${b}_{0}\in R^{d} $ are the learnable parameters, $d_0$ is the initial feature dimension of $v_{i}$.

\subsubsection{Structural Self-Attention}
\label{sub:sa}


Note that traditional graph transformers attention projects query and key vectors from each node feature and then computes a full dot-product between them, representing the attention to which node $v_i$ attends to another node $v_j$ \cite{DBLP:conf/icml/ChenOB22,dwivedi2020generalization}.
That is, self-attention is agnostic to learning structural similarity and proximity between pairs of nodes, resulting in poor performance in graph structure preservation.
Therefore, we propose a structural self-attention to capture the proximity and structural similarity between node pairs.
Specifically, the attention score at the $l$-th layer of the $k$-th head can be formulated as:
\begin{eqnarray}
\label{eq:selfattention}
&{{\alpha}_{ij}^{k,l}}=\frac{\left (Q^{k,l}\left[ h_{i}^{l}, s^{l}_{ij}\right ]\right)\left ({{K}^{k,l}}\left[ h_{j}^{l}, s^{l}_{ij}\right ]\right )}{\sqrt{{{d}_{k}}}}+{{\Phi}^{k,l}} f_d ^l\left( D_{ij}\right), \\ 
& Q^{k,l}\left[ h_{i}^{l}, s^{l}_{ij}\right ] =\left[ W_{n}^{Q^{k,l}}h_{i}^{l}+W_{s}^{Q^{k,l}}s^{l}_{ij}\right] , \\ 
& K^{k,l}\left[ h_{j}^{l}, s^{l}_{ij}\right ] =\left[ W_{n}^{K^{k,l}}h_{j}^{l}+W_{s}^{K^{k,l}}s^{l}_{ij}\right] ,
\end{eqnarray}
where ${W}^{Q^{k,l}}_n$, ${W}^{Q^{k,l}}_s$, ${W}^{K^{k,l}}_n$,  ${W}^{K^{k,l}}_s$, and ${\Phi}^{^{k,l}} \in R^{d_{k}\times d}$ are learnable transformation matrices, 
$h_{i}^{l}$ and $h_{j}^{l}$ are the node feature vectors of $v_{i}$ and $v_{j}$ at the $l$-th layer, respectively, 
$s_{ij}$ denotes the high-order proximity vector between $v_i$ and $v_j$, and $f_d\left( D_{ij}\right)$ is the linearly transformed degree-weighted score $D_{ij}$ between $v_i$ and $v_j$. 
We use high-order proximity $s_{ij}$ and degree-weighted score $D_{ij}$  to reflect community membership and roles in communities, respectively.
We use two separate learnable transformation matrices for node feature vectors and high-order proximity vectors, ${W}^{^{k,l}}_n$ and  ${W}^{^{k,l}}_s$, to obtain query and key matrices.


To define the matrix $s$, we analyze higher-order proximity between nodes in a multi-scale manner to examine whether they are densely connected across $k$-hops. 
$s_{ij}$ represents the high-order proximity between $v_i$ and $v_j$ can be defined as:
\begin{eqnarray}
{{s}_{ij}}&=&f_s\left( sim_{ij}^{(1)},sim_{ij}^{(2)}, \cdots ,sim_{ij}^{(k)}\right) , \\ 
sim_{ij}^{(k)}&=&\frac{{{N}^{(k)}}(v_i)\cap{{N}^{(k)}}(v_j)}{{{N}^{(k)}}(v_i)\cup{{N}^{(k)}}(v_j)}\ ,
\end{eqnarray}
where $f_s(\cdot)$ represents a learnable linear transformation that maps a $k$-dimensional input vector to a $d$-dimensional output vector, 
${{N}^{(k)}}(v_i)$ refers to the set of nodes within $k$-hop distance rooted at node $v_i$, 
$s_{ij} \in R^{1 \times k}$ denotes the proximity score between two nodes $v_i$ and $v_j$, 
and $sim_{ij}^{(k)}$ denotes a measure of the proximity between two nodes at the $k$-th step.
The value of $sim_{ij}^{(k)}$ quantifies how close two nodes are in terms of the shared neighbors between two nodes $v_i$ and $v_j$ at the $k$-hop distance. 
This value captures the structural proximity of two nodes based on the number of shared neighbors at the $k$-hop distance.
Besides, the number of consecutive zeros from the first component of $sim_{ij}$ can tell the model about the $k$-hop distance between $v_i$ and $v_j$.
That is, the number of the first zeros in $sim_{ij}$ indicates how many hops are required before there is a transition probability between the two nodes.

\subsubsection{Graph Transformer Layers}
\label{Graph_Transformer_Layers}

We then aggregate messages from the context nodes using the attention scores and concatenate $H$ outputs of the aggregation on each head into an updated node feature vector, followed by a linear projection. 
The node feature vector $h^{l}_i$ of $v_i$ at the $l$-th layer is then updated as:
\begin{equation}
\label{eq:12}
\hat h_{i}^{l+1}=O_{h}^{l}\underset{k=1}{\overset{H}{\mathop{\mathbin\Big\Vert}}}\, \left(\sum\limits_{v_j\in N({{v}_{i}})}{\tilde{\alpha}_{ij}^{k,l}{{V}^{k,l}}h_{j}^{l}}\right),
\end{equation}
where ${\tilde{\alpha}_{ij}^{k,l}}=\text{softmax}_{j}({{\alpha}_{ij}^{k,l}})$,  
${V}^{k,l} \in R^{d_k \times d }$ indicates the value matrix, 
$O_{h}^{l} \in R^{d \times d }$ is a learnable transformation matrix, and
$\mathbin\Vert $ refers to concatenation. 
We then pass the outputs to feed-forward networks (FFN) by adding residual connections and layer normalization as:
\begin{align}
\hat{\hat{h}}_{i}^{l+1}& =\text{LN}\left( h_{i}^{l}+\hat{h}_{i}^{l+1}\right), \\
 h_{i}^{l+1} &= W_{2}^{l}\sigma\left( W_{1}^{l}\hat{\hat{h}}_{i}^{l+1}\right), \
\end{align}
where $W_{1}^{l}\in R^{2d\times d}$ and $W_{2}^{l}\in R^{d\times 2d}$ are learnable parameters, $\sigma(\cdot)$ denotes the ReLU function, and $\text{LN}(\cdot)$ indicates layer normalization.

\subsection{Self-Supervised Learning Tasks}
\label{subsec:ssl}




We propose an effective Self-Supervised Learning task for preserving the global graph structures without using any label information.
In this way, DegFairGT could learn fair embeddings to preserve the input graph structure, which then can be adapted to various downstream tasks.
To achieve this, DegFairGT explicitly preserves the graph structure as a graph reconstruction task by leveraging the $p$-step transition probability matrix. 
The transition matrix captures higher-order structural relationships beyond the adjacent neighbors, presenting both local and global connectivity, ensuring that embeddings preserve local neighborhoods and global structural information.

Given the output embeddings from the final transformer layer, we aim to reconstruct the $k$-step transition matrix probability of the input graph \cite{DBLP:journals/corr/abs-2308-09517,GutmannH12}.
Formally, given the output embeddings of any two nodes $v_i$ and $v_j$, as $Z_i$ and $Z_j$, respectively, we compute the similarity score between them, i.e., cosine similarity, as  $Z^*_{ij} = \frac{Z^{\top}_{i}{Z_j}}{\| {Z_i} \|\| {Z_j} \|}$.
In addition, we note that while the transition probability reconstruction ensures that embeddings retain the topological structure, node features play a crucial role in many real-world graphs.
That is, different graph domains, e.g., social networks and molecular graphs, also rely on different feature information, which can benefit the model in learning fair representations.
Therefore, in addition to learning graph structure, DegFairGT reconstructs the initial node features to enhance the adaptability of fair representations.
The loss function can be defined as:


\begin{equation}
 L_1 ={\beta}_1\sum\limits_{p}{  \left\|{M^{(p)}- Z^*}\right\|_{F}^{2}}+{\beta}_2\frac{1}{\left| V\right|}{{{\left\|{X}-{{{\hat{X}}}}\right\|}_{2}}}, 
\end{equation}
where $M^{(p)}$ is the pre-computed log-scaled $p$-step transition matrix from the input graph with NCE \cite{DBLP:journals/corr/abs-2308-09517}, $Z^*= Z\cdot Z^\text{T}$ is the cosine similarity matrix calculated from the output node representations $Z$, $X$ and $\hat{X}$ refer to the initial node features and the output node embeddings obtained by learnable linear transformation of $Z$, respectively, and $\beta_1$ and  $\beta_2$ are hyper-parameters.
$V$ denotes the set of nodes in the input graph.


In the early stages of training, the graph augmentation could collapse into insignificant edges and modify graph structures in an undesirable way. 
For example, the augmentation module could learn to produce a fully connected graph or excessively remove edges between adjacent nodes that retain no original graph structure. 
To prevent the structural augmentation from collapsing into insignificant edges, we use a binary cross entropy (BCE) loss to penalize the augmentation, as follows:
\begin{equation}
\resizebox{0.41\textwidth}{!}{ ${{L}_{2}}=-\sum\limits_{i,j=1}^{N}{\left[{{A}_{ij}}\log ( A_{ij}^{'})+\left( 1-{{A}_{ij}}\right)\log ( 1-A_{ij}^{'})\right]} $ }, 
\end{equation}
where $A$ is the original adjacency matrix of the input graph,
and $A'$ is the adjacency matrix of the augmented graph.
The total loss functions for the pre-training process are then combined as:
\begin{equation}
\label{eq:total_loss}
L = L_1 + \alpha L_2,
\end{equation}
where $\alpha$ is the hyper-parameter to control the balance between structure preservation and augmentation losses.


\section{Evaluation}

To demonstrate DegFairGT's ability to mitigate degree bias problems, we conducted a series of fairness analyses, which analyzed the model performance differences between high- and low-degree node groups for the node classification benchmarks.
Also, we compared DegFairGT with the state-of-the-art graph transformers and graph augmentation-based methods on the node classification task and node clustering task to evaluate the overall performance and capability of preserving graph structural features, respectively.
The source code is available in our code repository\footnote{https://github.com/NSLab-CUK/Community-aware-Graph-Transformer}.



\subsection{Experimental Settings}

\subsubsection{Datasets}

We considered six publicly available datasets grouped into three domains, including citation networks (Cora, Citeseer, and Pubmed~\cite{sen2008collective}), co-purchase networks (Amazon Computers and Photo~\cite{DBLP:conf/sigir/McAuleyTSH15}), and a reference network (WikiCS~\cite{DBLP:journals/corr/abs-2007-02901}).
The detailed datasets are summarized in Table \ref{tab:Benchmarks}. 
We conducted each experiment ten times by randomly sampling training, validation, and testing sets with sizes of 60\%, 20\%, and 20\%, respectively.
The table results show the mean and standard deviation of the metrics on the testing set over the ten cases.

\begin{table*}[tb]
\centering
\setlength{\tabcolsep}{6 pt}
\caption{A summary of statistics of datasets.
\textcolor{red}{
}
}
\begin{tabular}{lccccc}
\toprule
Dataset &\# Nodes & \# Edges & \# Classes &Feature Dimension & Density ($\times10^{-4}$)  \\

\midrule
Cora~\cite{sen2008collective}
&2,708
&5,429
&7
&1,433
&14.81
\\
Citeseer~\cite{sen2008collective}
&3,327
&4,732
&6
&3,703
&8.55 
\\
Pubmed~\cite{sen2008collective}
& 3,703
& 44,338
& 3
& 500
&64.68 
\\
Computers~\cite{DBLP:conf/sigir/McAuleyTSH15}
&  13,752
&  245,861
&  10
&  767
&26.00 
\\
Photo~\cite{DBLP:conf/sigir/McAuleyTSH15}
&7,650
&119,081
&8
&745
&40.70 
\\
WikiCS~\cite{DBLP:journals/corr/abs-2007-02901}
&11,701
&216,123
&10
&300
&31.57 
\\
\bottomrule
\end{tabular}
\label{tab:Benchmarks}
\end{table*}

\subsubsection{Baselines}


We compared DegFairGT to the three groups of baselines: popular GNN architectures, fairness, and augmentation-based methods, as well as graph transformers.
Our goal is to validate the performance of DegFairGT to address the degree bias and graph structure preservation compared to recent studies.
For baseline settings, we follow closely from these studies for fair comparison.
For popular GNN architectures, we considered two methods, i.e., GIN and GAT:
\begin{itemize}
    \item GIN \cite{DBLP:conf/iclr/XuHLJ19} is an expressive GNN to distinguish between non-isomorphic graph structures, which enables the model to generate well-distinguished representations. 
    \item GAT \cite{velivckovic2017graph} uses an attention mechanism to capture the importance of each neighboring node during message-passing aggregation.
\end{itemize}
For fairness and augmentation-based methods, we considered five recent methods:
\begin{itemize}
    \item GRACE \cite{Zhu:2020vf} generates multiple views of graphs by randomly removing edges and masking node features to adjust the edges in graphs. 
    The model then uses a contrastive objective to maximize the agreement between different views.
    \item GCA \cite{DBLP:conf/www/0001XYLWW21} generates informative edges by focusing on important structures via an adaptive augmentation with a bias on node centrality.
    \item RGRL \cite{DBLP:conf/cikm/LeeH0P22} randomly generates augmented graphs with a bias for low-degree nodes and then employs GNNs with contrastive objectives to learn representations.
    \item DegFairGNN \cite{DBLP:conf/aaai/LiuN023} aims to modulate the messages at each GNN layer to distill or complement information for high- and low-degree nodes based on debiasing message aggregation. 
    \item GRADE \cite{DBLP:conf/nips/Wang0SS22} samples context nodes from substructures surrounding each get node to generate informative edges with a low-degree node bias.
\end{itemize}
For graph transformers, we considered seven methods that can capture the global graph structures to mitigate the degree bias problem:
\begin{itemize}
    \item GT \cite{dwivedi2020generalization} is a vanilla graph transformer that learns the representations via self-attention and employs a Laplacian positional encoding as node features.
    \item SAN \cite{DBLP:conf/nips/KreuzerBHLT21} aims to adaptively capture the graph structures by using a learnable positional encoding.
    \item SAT \cite{DBLP:conf/icml/ChenOB22} extracts $k$-hop subgraphs to aggregate messages for each target node based on GNNs followed by graph transformers.
    \item ANS\_GT \cite{DBLP:conf/nips/Zhang0HL22} construct context nodes by employing different sampling strategies, i.e., $k$-hop neighborhoods, node features similarity or node centrality, to capture the global information for each node.
    \item Graphormer \cite{DBLP:conf/nips/YingCLZKHSL21} aims to capture $k$-hop subgraphs surrounding each target node based on the shortest path distance and then incorporate the structural distance into self-attention.
    \item NodeFormer \cite{DBLP:conf/nips/WuZLWY22} aims at message-passing modulation at each layer, which could adaptively sample sufficient messages from neighborhoods.
    \item UGT \cite{DBLP:journals/corr/abs-2308-09517} introduces the use of structural identity to explore nodes with high structural similarity and then generates edges connecting non-adjacent nodes.
\end{itemize}

%

\subsubsection{Evaluation Metrics}

To analyze degree bias, we adopt two measurements, Degree Statistical Parity (${{\Delta}_{SP}}$) and Degree Equal Opportunity (${{\Delta}_{EO}}$) \cite{DBLP:conf/aaai/LiuN023,DBLP:conf/icml/MadrasCPZ18}.
The two metrics consider whether the distribution of predicted labels of each node is consistent between node groups of different generalized degrees.
For the details of the generalized degree, we refer readers to \cite{DBLP:conf/aaai/LiuN023}.
Intuitively, from the test set, we form two node groups $G_1$ and $G_2$ with low and high generalized degrees within $r$-hop distance, respectively, and compute ${{\Delta }_{SP}}$ and ${{\Delta }_{EO}}$.
Specifically, these metrics assess whether the distribution of predicted labels of each node is consistent between node groups of different generalized degrees.
We then compute ${{\Delta }_{SP}}$ and ${{\Delta }_{EO}}$ as:
\begin{align}
{\Delta }_{SP}&=\frac{1}{|\mathbb{C}|}\sum\nolimits_{c\in \mathbb{C}}| P_1( {{{\hat{y}}}_{v}}=c  ) - P_2( {{{\hat{y}}}_{v}}=c  ) | , \\      
{\Delta }_{EO}&=\frac{1}{|\mathbb{C}|}\sum\nolimits_{c\in \mathbb{C}} | P_1 ( {{{\hat{y}}}_{v}}=c|{{y}_{v}}=c ) -  P_2( {{{\hat{y}}}_{v}}=c|{{y}_{v}}=c ) |, 
\end{align}
where $\mathbb{C} \ni c$ is the set of classes in the testing set, $\hat{y}_v$ and ${y}_v$ denote the predicted and actual classes of node $v$, respectively, and
$P_1(\cdot)$ and $P_2(\cdot)$ are the ratios of nodes classified into each class in node groups $G_1$ and $G_2$, respectively. 

We use accuracy to evaluate the performance of the node classification task.
For the node clustering task, we employ the conductance (C) and modularity (Q) to evaluate the quality of the clustering results \cite{DBLP:journals/kais/YangL15}.

\subsubsection{Model Training and Resources}
\label{app:Model_training} 

Detailed hyperparameter specifications are given in Table \ref{tab:hyperparameter}.
We conducted a search on the embedding dimensions $\{32, 64, 128, 256\}$.
The hyperparameters $\alpha_2$, $\beta_1$, and $\beta_2$ are determined with a grid search among $\{0.01, 0.1, 0.5, 1, 10 \}$.
For the pre-training phase, we train the model for 500 epochs using Adam optimizer with $1\times 10^{-4}$ learning rate and $1\times 10^{-5}$ weight decay.

\begin{table}[tb]
\caption{Hyperparameters used in experiments.
}
\centering
\setlength{\tabcolsep}{6pt}
\begin{tabular}{l c }
\toprule
Hyperparameters
&Values
\\ 
\midrule

Number of Transformer layers 
& 4
\\
Number of Heads 
& 4
\\
Number of epochs for pre-training
& 500
\\
Number of Clusters
& 5
\\
Dropout rate
& 0.1
\\
Embedding dimension
&64
\\ 
Adam: initial learning rate
&$1\times 10^{-4}$
\\
Adam: weight decay
& $1\times 10^{-5}$
\\
Activation function
& ReLU
\\

$\alpha$
&1.0
\\
$\beta_1$
&0.5
\\
$\beta_2$
&0.5
\\
$\xi$
&0.8
\\
$\zeta$ 
&0.2

\\
\bottomrule
\end{tabular}
\label{tab:hyperparameter}
\end{table}



We conducted the experiments in two servers, each equipped with four NVIDIA RTX A5000 GPUs and 24GB RAM/GPU.
DegFairGT was built and tested in Python 3.8.8 using Torch-geometric Library \cite{DBLP:journals/corr/abs-1903-02428} and DGL Library \cite{wang2019dgl}.
The experiments were tested on the Ubuntu 20.04 LTS.

\subsection{Fairness Analysis}

\begin{table*}[t]
\centering
\setlength{\tabcolsep}{2.5 pt} 
\caption{A fairness analysis on $r=1$ and Top/Bottom 20\%.
The top two are highlighted by \textbf{first} and \underline{second}.
} 
\begin{tabular}{l cc cc cc cc cc cc}
   \toprule
\multirow{1}{*}{}
        &\multicolumn{2}{c}{Cora} 
        &\multicolumn{2}{c}{Citeseer} 
        &\multicolumn{2}{c}{Pubmed} 
        & \multicolumn{2}{c}{Computers}
        &\multicolumn{2}{c}{Photo}
        &\multicolumn{2}{c}{WikiCS}\\
        
       \cmidrule(lr){2-3}
       \cmidrule(lr){4-5}
       \cmidrule(lr){6-7}
       \cmidrule(lr){8-9}
       \cmidrule(lr){10-11}
       \cmidrule(lr){12-13}

& ${{\Delta}_{SP}}\downarrow $
&  ${{\Delta}_{EO}}\downarrow $

& ${{\Delta}_{SP}}\downarrow $
&  ${{\Delta}_{EO}}\downarrow $

& ${{\Delta}_{SP}}\downarrow $
&  ${{\Delta}_{EO}}\downarrow $

& ${{\Delta}_{SP}}\downarrow $
&  ${{\Delta}_{EO}}\downarrow $

& ${{\Delta}_{SP}}\downarrow $
&  ${{\Delta}_{EO}}\downarrow $

& ${{\Delta}_{SP}}\downarrow $
&  ${{\Delta}_{EO}}\downarrow $

\\\midrule
GIN
&4.77\tiny{$\pm$0.68}
&19.82\tiny{$\pm$4.37}

&9.55\tiny{$\pm$0.30}
&19.80\tiny{$\pm$4.37}

&7.99\tiny{$\pm$0.72 }
&10.45\tiny{$\pm$1.27}

&13.39\tiny{$\pm$1.02}
&34.05\tiny{$\pm$6.01}

&12.55\tiny{$\pm$0.81}
&14.72\tiny{$\pm$2.21}

&10.23\tiny{$\pm$0.35}
&34.30\tiny{$\pm$2.68}
\\

GAT
&5.20\tiny{$\pm$1.16}
&12.46\tiny{$\pm$1.43}

&9.78\tiny{$\pm$1.22}
&17.21\tiny{$\pm$5.64}

&7.20\tiny{$\pm$1.07}
&12.04\tiny{$\pm$0.64}

&{5.42\tiny{$\pm$1.18}}
&18.33\tiny{$\pm$1.42}

&\underline{\text{8.35\tiny{$\pm$0.87}}}
&11.52\tiny{$\pm$2.07}

&8.74\tiny{$\pm$0.93}
&10.96\tiny{$\pm$1.15}
\\
\midrule
RGRL
&5.83\tiny{$\pm$0.74}
&13.90\tiny{$\pm$2.16}

&8.86\tiny{$\pm$0.89}
&13.32\tiny{$\pm$2.59}

&8.73\tiny{$\pm$1.59}
&13.32\tiny{$\pm$1.10}

&5.92\tiny{$\pm$0.27}
&16.02\tiny{$\pm$2.18}

&9.52\tiny{$\pm$0.41}
&20.53\tiny{$\pm$1.56}
&8.28\tiny{$\pm$0.34}
&16.83\tiny{$\pm$2.27}
\\

{GRACE}
&4.89\tiny{$\pm$0.24}
&{8.90\tiny{$\pm$2.79}}

&8.94\tiny{$\pm$0.96}
&\underline{\text{13.20\tiny{$\pm$0.99}}}

&8.57\tiny{$\pm$0.55}
&11.62\tiny{$\pm$0.26}

&6.24\tiny{$\pm$0.24}
&13.10\tiny{$\pm$0.85}

&9.68\tiny{$\pm$0.10}
&14.57\tiny{$\pm$0.27}
&10.34\tiny{$\pm$0.83}
&17.33\tiny{$\pm$1.45}
\\
GCA
&4.51\tiny{$\pm$0.09}
&10.39\tiny{$\pm$0.83}

&8.75\tiny{$\pm$0.47}
&14.57\tiny{$\pm$2.70}

&8.17\tiny{$\pm$0.34}
&14.25\tiny{$\pm$2.34}

&5.87\tiny{$\pm$0.25}
&16.07\tiny{$\pm$1.06}

&9.59\tiny{$\pm$0.16}
&22.61\tiny{$\pm$0.18}
&8.29\tiny{$\pm$0.67}
&17.06\tiny{$\pm$2.52}
\\
DegFairGNN 
&\underline{\text{4.01\tiny{$\pm$0.66}}}
&14.86\tiny{$\pm$2.82}

& 13.85\tiny{$\pm$2.80}
&18.79\tiny{$\pm$4.99}

&6.24\tiny{$\pm$0.47 }
&7.20\tiny{$\pm$0.13 }

&5.76\tiny{$\pm$0.06 }
&15.81\tiny{$\pm$1.52 }

&10.39\tiny{$\pm$0.12 }
&18.76\tiny{$\pm$1.14 }

&9.01\tiny{$\pm$1.85 }
&18.76\tiny{$\pm$2.55}
\\
GRADE
&4.75\tiny{$\pm$0.86}
&12.67\tiny{$\pm$2.05}

&8.74\tiny{$\pm$1.54}
&14.49\tiny{$\pm$2.48}

&8.23\tiny{$\pm$1.24}
&13.85\tiny{$\pm$2.14}

&6.06\tiny{$\pm$1.72}
&\textbf{\text{11.33\tiny{$\pm$1.93}}}

&12.70\tiny{$\pm$2.77 }
&12.86\tiny{$\pm$2.16}

&8.76\tiny{$\pm$2.02 }
&{10.62\tiny{$\pm$1.96}}
\\






\midrule
GT
&5.86\tiny{$\pm$1.00}
&14.13\tiny{$\pm$1.51}

&8.34\tiny{$\pm$1.99}
&16.31\tiny{$\pm$0.23}

&5.12\tiny{$\pm$0.41}
&\underline{\text{6.15\tiny{$\pm$1.06}}}

&5.69\tiny{$\pm$0.03}
&16.40\tiny{$\pm$2.81}

&9.25\tiny{$\pm$0.76}
&\textbf{\text{10.59\tiny{$\pm$2.67}}}

&{7.94\tiny{$\pm$0.15}}
&12.96\tiny{$\pm$3.24}
\\
SAN
&4.90\tiny{$\pm$0.82}
&10.00\tiny{$\pm$0.40}

&10.80\tiny{$\pm$1.60}
&17.44\tiny{$\pm$3.03}

&\underline{\text{4.09\tiny{$\pm$1.75}}}
&8.30\tiny{$\pm$0.92}

&5.43\tiny{$\pm$0.70}
&17.48\tiny{$\pm$4.19}

&9.24\tiny{$\pm$0.04}
&13.68\tiny{$\pm$1.18}
&\underline{\text{7.73\tiny{$\pm$0.28}}}
&16.23\tiny{$\pm$1.07}
\\
SAT
&4.76\tiny{$\pm$0.52}
&14.91\tiny{$\pm$1.28}

&7.13\tiny{$\pm$1.84}
&14.66\tiny{$\pm$1.21}

&4.57\tiny{$\pm$0.90}
&7.54\tiny{$\pm$1.97}

&5.97\tiny{$\pm$0.58}
&20.26\tiny{$\pm$1.75}

&{8.94\tiny{$\pm$1.80}}
&15.44\tiny{$\pm$1.27}
&8.83\tiny{$\pm$1.76}
&14.31\tiny{$\pm$2.55}
\\
ANS\_GT
&{4.17\tiny{$\pm$0.69}}
&13.71\tiny{$\pm$1.93}

&\underline{\text{6.46\tiny{$\pm$1.31}}}
&{13.27\tiny{$\pm$1.35}}

&{4.46\tiny{$\pm$0.97}}
&14.41\tiny{$\pm$0.85}

&\textbf{\text{4.90\tiny{$\pm$0.85}}}
&\underline{\text{12.03\tiny{$\pm$1.43}}}

&8.74\tiny{$\pm$0.38}
&12.51\tiny{$\pm$2.12}

&8.23\tiny{$\pm$0.35}
&14.30\tiny{$\pm$2.68}
\\ 
Graphormer
&4.36\tiny{$\pm$0.22}
&12.30\tiny{$\pm$1.69}

&{6.82\tiny{$\pm$0.64}}
&13.40\tiny{$\pm$1.51}

&5.32\tiny{$\pm$0.44}
&7.88\tiny{$\pm$0.81}

&5.96\tiny{$\pm$1.56}
&12.17\tiny{$\pm$1.20}

&8.96\tiny{$\pm$0.24}
&{11.26\tiny{$\pm$1.77}}

&8.07\tiny{$\pm$0.33}
&15.79\tiny{$\pm$0.41}
\\
NodeFormer
&4.38\tiny{$\pm$0.42}
&\underline{\text{7.82\tiny{$\pm$1.72}}}

&9.39\tiny{$\pm$1.76}
&14.15\tiny{$\pm$2.81}

&5.52\tiny{$\pm$1.21}
&11.40\tiny{$\pm$1.52}

& 5.88\tiny{$\pm$0.47}
&12.50\tiny{$\pm$1.96}

&12.60\tiny{$\pm$2.28}
&14.49\tiny{$\pm$2.09}

&9.01\tiny{$\pm$1.62}
&\underline{\text{10.39\tiny{$\pm$1.88}}}
\\
UGT 
&4.32\tiny{$\pm$0.12}
&10.54\tiny{$\pm$.25}

&8.82\tiny{$\pm$0.30}
&15.57\tiny{$\pm$4.58}

&5.20\tiny{$\pm$0.14}
&{7.11\tiny{$\pm$0.62}}

&8.67\tiny{$\pm$0.08}
&16.22\tiny{$\pm$0.94}

&9.41\tiny{$\pm$0.25}
&14.31\tiny{$\pm$0.39}

&9.41\tiny{$\pm$0.25}
&14.31\tiny{$\pm$0.39}

\\
\midrule
\text{DegFairGT}
&\textbf{\text{3.46\tiny{$\pm$0.97}}}
&\textbf{\text{4.26\tiny{$\pm$0.24}}}

&\textbf{\text{5.93\tiny{$\pm$1.70}}}
&\textbf{\text{11.80\tiny{$\pm$3.81}}}

&\textbf{\text{3.29\tiny{$\pm$0.80}}}
&\textbf{\text{6.04\tiny{$\pm$0.52}}}

&\underline{\text{4.99\tiny{$\pm$0.64}}}
&{12.15\tiny{$\pm$0.84}}

&\textbf{\text{8.26\tiny{$\pm$0.18}}}
&\underline{\text{11.13\tiny{$\pm$3.05}}}

&\textbf{\text{6.41\tiny{$\pm$0.39}}}
&\textbf{\text{9.30\tiny{$\pm$1.19}}}

\\
\bottomrule
\end{tabular}
\label{tab:fainess_1_20}
\end{table*}

\begin{table*}
\centering
\setlength{\tabcolsep}{2.5 pt} 
\caption{A fairness analysis of DegFairGT and the baselines on $r=2$ and Top/Bottom 20\%. 
}
\begin{tabular}{l cc cc cc cc cc cc}
   \toprule
\multirow{1}{*}{}
        &\multicolumn{2}{c}{Cora} 
        &\multicolumn{2}{c}{Citeseer} 
        &\multicolumn{2}{c}{Pubmed} 
        & \multicolumn{2}{c}{Computers}
        &\multicolumn{2}{c}{Photo}
        &\multicolumn{2}{c}{WikiCS}\\
        
       \cmidrule(lr){2-3}
       \cmidrule(lr){4-5}
       \cmidrule(lr){6-7}
       \cmidrule(lr){8-9}
       \cmidrule(lr){10-11}
       \cmidrule(lr){12-13}

& ${{\Delta}_{SP}}\downarrow $
&  ${{\Delta}_{EO}}\downarrow $

& ${{\Delta}_{SP}}\downarrow $
&  ${{\Delta}_{EO}}\downarrow $

& ${{\Delta}_{SP}}\downarrow $
&  ${{\Delta}_{EO}}\downarrow $

& ${{\Delta}_{SP}}\downarrow $
&  ${{\Delta}_{EO}}\downarrow $

& ${{\Delta}_{SP}}\downarrow $
&  ${{\Delta}_{EO}}\downarrow $

& ${{\Delta}_{SP}}\downarrow $
&  ${{\Delta}_{EO}}\downarrow $

\\\midrule
GIN
&10.14\tiny{$\pm$1.72}
&34.45\tiny{$\pm$2.72}

&15.21\tiny{$\pm$1.41}
&31.01\tiny{$\pm$1.12}

&\textbf{\text{10.20\tiny{$\pm$1.31}}}
&\textbf{\text{8.23\tiny{$\pm$4.17}}}

&17.11\tiny{$\pm$0.35}
&30.83\tiny{$\pm$4.91}

&17.89\tiny{$\pm$1.06}
&27.62\tiny{$\pm$4.76}

&13.10\tiny{$\pm$0.68}
&41.07\tiny{$\pm$2.88}
\\

GAT
&10.36\tiny{$\pm$0.59}
&23.86\tiny{$\pm$5.28}

&15.85\tiny{$\pm$1.30}
&25.36\tiny{$\pm$5.15}

&15.21\tiny{$\pm$1.47}
&22.76\tiny{$\pm$1.66}

&10.99\tiny{$\pm$1.53}
&14.03\tiny{$\pm$2.49}

&15.19\tiny{$\pm$0.72}
&17.19\tiny{$\pm$4.90}

&10.68\tiny{$\pm$0.58}
&11.09\tiny{$\pm$0.94}
\\
\midrule
RGRL
&{9.19\tiny{$\pm$0.91}}
&21.77\tiny{$\pm$3.05}

&15.40\tiny{$\pm$1.38}
&19.26\tiny{$\pm$2.70}

&16.75\tiny{$\pm$1.33}
&22.60\tiny{$\pm$1.23}

&11.39\tiny{$\pm$0.23}
&18.26\tiny{$\pm$2.85}

&14.45\tiny{$\pm$0.40}
&24.11\tiny{$\pm$2.60}

&10.19\tiny{$\pm$0.40}
&19.28\tiny{$\pm$2.63}
\\

{GRACE}
&9.20\tiny{$\pm$0.08}
&17.71\tiny{$\pm$4.18}

&15.25\tiny{$\pm$0.83}
&18.52\tiny{$\pm$4.09}

&11.62\tiny{$\pm$0.26}
&16.77\tiny{$\pm$0.85}

& 9.78\tiny{$\pm$1.81}
&13.10\tiny{$\pm$0.85}

&15.73\tiny{$\pm$0.39}
&19.47\tiny{$\pm$1.41}

&11.46\tiny{$\pm$0.68}
&18.80\tiny{$\pm$1.46}
\\
GCA
&9.48\tiny{$\pm$0.73}
&20.45\tiny{$\pm$7.16}

&15.48\tiny{$\pm$0.49}
&16.01\tiny{$\pm$0.55}

&18.43\tiny{$\pm$1.07}
&23.13\tiny{$\pm$0.76}

&12.44\tiny{$\pm$0.06}
&13.31\tiny{$\pm$0.12}

&15.71\tiny{$\pm$0.14}
&25.83\tiny{$\pm$0.70}

&11.26\tiny{$\pm$0.34}
&19.70\tiny{$\pm$3.09}
\\
DegFairGNN
&9.64\tiny{$\pm$0.72 }
&19.87\tiny{$\pm$2.37 }

&15.16\tiny{$\pm$2.51 }
&16.25\tiny{$\pm$4.73 }

&13.87\tiny{$\pm$0.76}
&17.76\tiny{$\pm$0.57}

&12.33\tiny{$\pm$0.10}
&12.63\tiny{$\pm$0.89}

&16.80\tiny{$\pm$0.16}
&29.40\tiny{$\pm$1.90}

&10.16\tiny{$\pm$2.05}
&18.66\tiny{$\pm$7.22}
\\

GRADE
&11.56\tiny{$\pm$1.33  }
&17.68\tiny{$\pm$3.42}

&14.98\tiny{$\pm$3.39}
&20.86\tiny{$\pm$2.47}

&14.36\tiny{$\pm$2.24}
&24.74\tiny{$\pm$4.44}

&{8.96\tiny{$\pm$2.01}}
&15.11\tiny{$\pm$3.69}

&{13.55\tiny{$\pm$4.31}}
&14.18\tiny{$\pm$3.02}

&14.22\tiny{$\pm$2.92}
&17.46\tiny{$\pm$3.23}

\\

\midrule
GT
&9.72\tiny{$\pm$1.37}
&18.19\tiny{$\pm$2.87}

&13.47\tiny{$\pm$2.35}
&17.29\tiny{$\pm$3.97}

&{10.81\tiny{$\pm$1.30}}
&{11.72\tiny{$\pm$2.50}}

&10.58\tiny{$\pm$0.38}
&12.81\tiny{$\pm$1.37}

&15.87\tiny{$\pm$0.80}
&11.61\tiny{$\pm$2.37}

&10.13\tiny{$\pm$0.27}
&12.23\tiny{$\pm$2.96}
\\
SAN
&10.30\tiny{$\pm$1.29}
&{15.55\tiny{$\pm$1.44}}

&16.68\tiny{$\pm$0.96}
&20.01\tiny{$\pm$5.63}

&12.24\tiny{$\pm$2.04}
&15.56\tiny{$\pm$1.27}

&10.27\tiny{$\pm$0.36}
&12.83\tiny{$\pm$2.08}

&16.18\tiny{$\pm$0.01}
&19.76\tiny{$\pm$1.38}

&9.87\tiny{$\pm$0.56}
&22.36\tiny{$\pm$5.31}

\\
SAT
&9.41\tiny{$\pm$0.77}
&16.57\tiny{$\pm$1.14}

&10.45\tiny{$\pm$0.92}
&{13.17\tiny{$\pm$1.99}}

&11.55\tiny{$\pm$1.25}
&12.35\tiny{$\pm$1.65}

&9.52\tiny{$\pm$0.73}
&13.29\tiny{$\pm$0.69}

&15.84\tiny{$\pm$0.97}
&12.79\tiny{$\pm$1.82}

&10.22\tiny{$\pm$0.46}
&14.11\tiny{$\pm$1.54}
\\
ANS\_GT
&13.18\tiny{$\pm$1.19}
&17.65\tiny{$\pm$2.34}

&{10.13\tiny{$\pm$1.37}}
&14.05\tiny{$\pm$1.91}

&11.39\tiny{$\pm$0.58}
&15.84\tiny{$\pm$1.04}

&\underline{\text{8.18\tiny{$\pm$1.19}}}
&14.65\tiny{$\pm$1.34}

&14.89\tiny{$\pm$1.61}
& 15.77\tiny{$\pm$2.17}

&{9.34\tiny{$\pm$1.99}}
&11.30\tiny{$\pm$2.02}
\\ 
Graphormer
&9.74\tiny{$\pm$0.55}
&\underline{\text{13.21\tiny{$\pm$1.70}}}

&\underline{\text{9.27\tiny{$\pm$1.21}}}
& \underline{\text{12.47\tiny{$\pm$2.24}}}

&11.53\tiny{$\pm$1.03}
&14.20\tiny{$\pm$2.23}

&10.76\tiny{$\pm$173}
&17.45\tiny{$\pm$3.52}

&13.96\tiny{$\pm$1.08}
&{11.26\tiny{$\pm$1.77}}

& 10.32\tiny{$\pm$0.34}
&19.72\tiny{$\pm$2.75}
\\
NodeFormer
&10.15\tiny{$\pm$1.69 }
&19.47\tiny{$\pm$2.27}

&10.25\tiny{$\pm$2.19}
&14.38\tiny{$\pm$3.81}

&10.88\tiny{$\pm$1.57}
&13.37\tiny{$\pm$3.04}

&11.39\tiny{$\pm$2.03}
&{11.46\tiny{$\pm$1.12}}

&\textbf{\text{12.38\tiny{$\pm$2.47}}}
&\underline{\text{10.45\tiny{$\pm$1.94}}}

&\textbf{\text{8.76\tiny{$\pm$2.17}}}
&{9.04\tiny{$\pm$2.65}}
\\
UGT 
&\underline{\text{9.16\tiny{$\pm$0.79}}}
&18.78\tiny{$\pm$2.83}

&10.43\tiny{$\pm$2.34}
&20.69\tiny{$\pm$2.39}

&22.53\tiny{$\pm$3.14}
&25.81\tiny{$\pm$2.40}

&11.10\tiny{$\pm$0.18}
&\underline{\text{7.16\tiny{$\pm$0.36}}}

&15.52\tiny{$\pm$0.18}
&20.81\tiny{$\pm$2.21}

&10.53\tiny{$\pm$0.14}
&\textbf{\text{6.41\tiny{$\pm$0.09}}}
\\
\midrule
\text{DegFairGT}
&\textbf{\text{8.71\tiny{$\pm$0.42}}}
&\textbf{\text{8.72\tiny{$\pm$0.73}}}

&\textbf{\text{8.86\tiny{$\pm$0.44}}}
&\textbf{\text{7.54\tiny{$\pm$1.47}}}

&\underline{\text{10.35\tiny{$\pm$9.48}}}
& \underline{\text{8.51\tiny{$\pm$1.17}}}

&\textbf{\text{5.18\tiny{$\pm$0.54}}}
&\textbf{\text{5.07\tiny{$\pm$0.85}}}

&\underline{\text{13.45\tiny{$\pm$1.47}}}
&\textbf{\text{7.43\tiny{$\pm$0.67}}}

&\underline{\text{9.08\tiny{$\pm$1.01}}}
&\underline{\text{8.48\tiny{$\pm$0.93}}}

\\
\bottomrule
\end{tabular}
\label{tab:fainess_2_20}
\end{table*}

We first evaluate DegFairGT against the baselines in terms of $\Delta_{SP}$ and $\Delta_{EO}$, with $r=1$ and 20\% Top/Bottom (i.e., $G_1$ and $G_2$ include the top and bottom 20\% of nodes in terms of the number of one-hop neighborhoods), as shown in Table \ref{tab:fainess_1_20}.
We have the following observations:
\textbf{(i)} DegFairGT consistently outperformed most baselines, e.g., GT, UGT, GRADE, and SAN, on both fairness metrics, showing the effectiveness of structural augmentation and structural self-attention.
For example, DegFairGT achieved the best performance on four datasets: Cora, Citeseer, Pubmed, and WikiCS on both $\Delta_{SP}$ and $\Delta_{EO}$ metrics and gained second and third ahead of other baselines on Computers and Photo dataset.
The improvement over the graph transformers could be caused by that $k$-hop neighborhood sampling can include noisy edges, and structural similarity can effectively refine the sampled context nodes.
\textbf{(ii)} Graph transformers, e.g., ANS\_GT and Graphormer, outperformed augmentation-based methods in most datasets regarding both fairness metrics.
This result indicates that aggregating $k$-hop neighborhoods, which guarantee at least a certain degree of proximity to target nodes, can contribute to addressing degree bias more than looking over a variety of nodes. 
We assume that random or heuristic augmentation-based methods, e.g., RGRL and GCA, can not eliminate possibilities enough for generating or leaving noisy edges connecting distant nodes with low proximity, which could hinder learning accurate representations. 
To this end, the superior performance of DegFairGT compared to baselines demonstrates that considering both node proximity and structural similarity enables our augmentation and self-attention to address degree bias by capturing correlated nodes.

\begin{table*}[tb]
\centering
\setlength{\tabcolsep}{2.5 pt} 
\caption{A fairness analysis on $r=1$ and Top/Bottom 30\%.
}
\begin{tabular}{l cc cc cc cc cc cc}
   \toprule
\multirow{1}{*}{}
&\multicolumn{2}{c}{Cora} 
&\multicolumn{2}{c}{Citeseer} 
&\multicolumn{2}{c}{Pubmed} 
& \multicolumn{2}{c}{Computers}
&\multicolumn{2}{c}{Photo}
&\multicolumn{2}{c}{WikiCS}
\\

\cmidrule(lr){2-3}
\cmidrule(lr){4-5}
\cmidrule(lr){6-7}
\cmidrule(lr){8-9}
\cmidrule(lr){10-11}
\cmidrule(lr){12-13}

& ${{\Delta}_{SP}}\downarrow $
&  ${{\Delta}_{EO}}\downarrow $

& ${{\Delta}_{SP}}\downarrow $
&  ${{\Delta}_{EO}}\downarrow $

& ${{\Delta}_{SP}}\downarrow $
&  ${{\Delta}_{EO}}\downarrow $

& ${{\Delta}_{SP}}\downarrow $
&  ${{\Delta}_{EO}}\downarrow $

& ${{\Delta}_{SP}}\downarrow $
&  ${{\Delta}_{EO}}\downarrow $

& ${{\Delta}_{SP}}\downarrow$
&  ${{\Delta}_{EO}}\downarrow$

\\\midrule

GIN
&4.28\tiny{$\pm$0.80}
&{7.02\tiny{$\pm$0.23}}

&\textbf{\text{4.71\tiny{$\pm$0.53}}}
&\underline{\text{7.93\tiny{$\pm$0.95}}}

&3.98\tiny{$\pm$0.03}
&7.11\tiny{$\pm$0.24}

&11.56\tiny{$\pm$0.80}
&33.85\tiny{$\pm$0.85}

&8.15\tiny{$\pm$0.63}
&24.80\tiny{$\pm$2.66}

&8.15\tiny{$\pm$1.56}
&30.19\tiny{$\pm$4.95}
\\

GAT
&4.71\tiny{$\pm$0.53}
&7.93\tiny{$\pm$0.95}

&10.38\tiny{$\pm$2.95}
&19.86\tiny{$\pm$3.94}

&6.80\tiny{$\pm$0.85}
&11.60\tiny{$\pm$3.14}

&4.71\tiny{$\pm$1.18}
&12.38\tiny{$\pm$2.05}

&7.10\tiny{$\pm$1.19}
&13.23\tiny{$\pm$3.21}

&6.70\tiny{$\pm$1.33}
&{9.26\tiny{$\pm$2.78}}
\\
\midrule
RGRL
&{3.52\tiny{$\pm$0.33}}
&7.64\tiny{$\pm$1.59}

&7.79\tiny{$\pm$0.87}
&12.71\tiny{$\pm$2.07}

&4.55\tiny{$\pm$1.12}
&9.02\tiny{$\pm$0.80}

&4.98\tiny{$\pm$0.73}
&9.99\tiny{$\pm$2.54}

&\textbf{\text{6.19\tiny{$\pm$0.28}}}
&13.73\tiny{$\pm$1.23}

&6.51\tiny{$\pm$0.30}
&14.25\tiny{$\pm$1.90}
\\

GRACE
&4.32\tiny{$\pm$0.40}
&9.47\tiny{$\pm$2.39}

&7.38\tiny{$\pm$1.43}
&10.61\tiny{$\pm$1.23}

&{3.61\tiny{$\pm$0.35}}
&7.24\tiny{$\pm$0.45}

&5.06\tiny{$\pm$0.10}
&11.09\tiny{$\pm$0.11}

&7.42\tiny{$\pm$0.23}
&11.38\tiny{$\pm$0.26}

&9.09\tiny{$\pm$0.55}
&15.57\tiny{$\pm$0.67}
\\
GCA
&3.62\tiny{$\pm$0.12}
&9.15\tiny{$\pm$0.99}

&7.19\tiny{$\pm$0.55}
&12.73\tiny{$\pm$0.80}

&\textbf{\text{3.06\tiny{$\pm$0.05}}}
&7.86\tiny{$\pm$0.08}

&4.97\tiny{$\pm$0.12}
&17.37\tiny{$\pm$0.17}

&7.42\tiny{$\pm$0.38}
&16.91\tiny{$\pm$3.96}

&13.27\tiny{$\pm$0.21}
&18.66\tiny{$\pm$1.38}
\\
DegFairGNN

&\underline{\text{3.48\tiny{$\pm$0.16}}}
&\underline{\text{6.69\tiny{$\pm$0.43}}}

&8.69\tiny{$\pm$0.28}
&11.43\tiny{$\pm$0.73}

&3.91\tiny{$\pm$0.51}
&{6.50\tiny{$\pm$1.00}}

&5.32\tiny{$\pm$0.16}
&10.35\tiny{$\pm$1.04}

&7.93\tiny{$\pm$0.19}
&13.18\tiny{$\pm$1.95}

&8.22\tiny{$\pm$0.78}
&14.63\tiny{$\pm$1.10}
\\ 
GRADE
&4.43\tiny{$\pm$1.14}
&12.23\tiny{$\pm$2.96}

&7.39\tiny{$\pm$0.83}
&16.99\tiny{$\pm$1.94}

&3.48\tiny{$\pm$0.86}
&6.58\tiny{$\pm$1.84}

&11.36\tiny{$\pm$1.35}
&10.15\tiny{$\pm$1.43}

&11.77\tiny{$\pm$1.89}
&9.66\tiny{$\pm$1.96}

&8.11\tiny{$\pm$1.39}
&9.71\tiny{$\pm$1.65}
\\ 

\midrule
GT
&4.71\tiny{$\pm$0.38}
&10.76\tiny{$\pm$1.78}

&8.76\tiny{$\pm$1.35}
&13.51\tiny{$\pm$2.04}

&3.72\tiny{$\pm$0.98}
&{6.32\tiny{$\pm$1.71}}

&4.92\tiny{$\pm$0.71}
&11.19\tiny{$\pm$2.61}

&6.66\tiny{$\pm$0.43}
&10.85\tiny{$\pm$1.18}

&6.88\tiny{$\pm$0.79}
&11.99\tiny{$\pm$1.81}
\\
SAN
&4.76\tiny{$\pm$0.94}
&14.32\tiny{$\pm$1.81}

&\underline{\text{5.64\tiny{$\pm$0.44}}}
&10.57\tiny{$\pm$0.58}

&4.24\tiny{$\pm$0.41}
&8.87\tiny{$\pm$1.77}

&4.38\tiny{$\pm$0.52}
&12.93\tiny{$\pm$0.18}

&7.31\tiny{$\pm$0.87}
&13.00\tiny{$\pm$2.54}

&7.06\tiny{$\pm$0.66}
&14.44\tiny{$\pm$2.65}
\\
SAT
&4.37\tiny{$\pm$0.78}
&17.59\tiny{$\pm$0.85}

&7.33\tiny{$\pm$0.18}
&16.14\tiny{$\pm$2.29}

&4.78\tiny{$\pm$0.41}
&\underline{\text{6.22\tiny{$\pm$1.48}}}

&4.57\tiny{$\pm$0.30}
&19.34\tiny{$\pm$2.62}

&6.77\tiny{$\pm$0.26}
&12.56\tiny{$\pm$2.09}

&7.01\tiny{$\pm$0.43}
&15.30\tiny{$\pm$1.87}

\\
ANS\_GT
&4.11\tiny{$\pm$1.15}
&12.72\tiny{$\pm$1.98}

&6.42\tiny{$\pm$1.41}
&13.06\tiny{$\pm$2.12}

&4.43\tiny{$\pm$0.23}
&7.18\tiny{$\pm$1.25}

&{4.22\tiny{$\pm$1.05}}
&9.81\tiny{$\pm$1.67}

&\underline{\text{6.46\tiny{$\pm$1.05}}}
&11.00\tiny{$\pm$3.74}

&{6.37\tiny{$\pm$1.91}}
&15.84\tiny{$\pm$3.24}

\\ 
Graphormer
&4.36\tiny{$\pm$0.22}
&12.30\tiny{$\pm$1.69}

&7.85\tiny{$\pm$1.53}
&{9.98\tiny{$\pm$2.36}}

&7.80\tiny{$\pm$0.52}
&10.03\tiny{$\pm$1.26}

&5.24\tiny{$\pm$0.99}
&9.54\tiny{$\pm$1.89}

&6.66\tiny{$\pm$1.49}
&{{6.82\tiny{$\pm$1.15}}}

&6.97\tiny{$\pm$1.82}
&12.90\tiny{$\pm$2.76}
\\
NodeFormer
&4.32\tiny{$\pm$1.18}
&13.79\tiny{$\pm$1.65}

&6.68\tiny{$\pm$1.44}
&11.83\tiny{$\pm$2.65}

&3.74\tiny{$\pm$1.68}
&6.56\tiny{$\pm$1.43}

&\underline{\text{4.19\tiny{$\pm$1.25}}}
&\underline{\text{8.28\tiny{$\pm$1.86}}}

&7.28\tiny{$\pm$1.13}
&15.74\tiny{$\pm$3.98}

&9.49\tiny{$\pm$1.18}
&9.65\tiny{$\pm$2.99}

\\
UGT 
&4.34\tiny{$\pm$0.01}
&11.61\tiny{$\pm$1.18}

&11.11\tiny{$\pm$0.07}
&15.66\tiny{$\pm$0.48}

&7.37\tiny{$\pm$0.79}
&13.57\tiny{$\pm$0.09}

&4.47\tiny{$\pm$0.02}
&\textbf{\text{3.87\tiny{$\pm$0.10}}}

&6.91\tiny{$\pm$0.22}
&\underline{\text{8.09\tiny{$\pm$1.34}}}

&\underline{\text{6.25\tiny{$\pm$0.01}}}
&\textbf{\text{7.95\tiny{$\pm$0.99}}}
\\
\midrule

\text{DegFairGT}
&\textbf{\text{3.38\tiny{$\pm$0.30}}}
&\textbf{\text{5.96\tiny{$\pm$0.78}}}

&{6.21\tiny{$\pm$0.84}}
&\textbf{\text{7.42\tiny{$\pm$0.03}}}

&\underline{\text{3.25\tiny{$\pm$0.17}}}
&\textbf{\text{6.04\tiny{$\pm$0.17}}}

&\textbf{\text{3.59\tiny{$\pm$0.38}}}
&{8.82\tiny{$\pm$0.68}}

&{6.59\tiny{$\pm$0.42}}
&\textbf{\text{6.05\tiny{$\pm$0.27}}}

&\textbf{\text{6.24\tiny{$\pm$0.47}}}
&\underline{\text{9.09\tiny{$\pm$0.81}}}
\\
\bottomrule
\end{tabular}
\label{tab:fainess_1_30}
\end{table*}

\begin{table*}[t]
\centering
\caption{
The performance on node classification.
}
\begin{tabular}{lcccccc}
\toprule
    &Cora &Citeseer &Pubmed  &Computers&Photo &WikiCS\\\midrule

 
  
GIN 
&77.25\tiny{$\pm$3.35}
&64.09\tiny{$\pm$1.95}
&85.96\tiny{$\pm$0.57}

&66.59\tiny{$\pm$0.16}
&88.92\tiny{$\pm$1.36}
&76.53\tiny{$\pm$0.82}

\\
GAT 
&84.21\tiny{$\pm$1.47}
&73.43\tiny{$\pm$1.21}
&82.43\tiny{$\pm$0.47}

&90.06\tiny{$\pm$0.76}
&93.34\tiny{$\pm$0.73}
&77.55\tiny{$\pm$0.71}
\\ 
\midrule
RGRL 
&84.27\tiny{$\pm$0.87}
&71.77\tiny{$\pm$0.89}
&82.50\tiny{$\pm$0.17}

&84.83\tiny{$\pm$0.43}
&92.14\tiny{$\pm$0.24}
&79.22\tiny{$\pm$0.49}
\\
GRACE 
&83.09\tiny{$\pm$0.86}
&69.28\tiny{$\pm$0.29}
&85.14\tiny{$\pm$0.24}

&88.12\tiny{$\pm$0.19}
&92.21\tiny{$\pm$0.10}
&30.52\tiny{$\pm$0.54}

\\
GCA 
&83.43\tiny{$\pm$0.34}
&68.20\tiny{$\pm$0.26}
&85.65\tiny{$\pm$1.02}

&74.87\tiny{$\pm$0.11}
&91.28\tiny{$\pm$0.58}
&32.34\tiny{$\pm$0.07}
\\
DegFairGNN 
&85.65\tiny{$\pm$1.52 }
&67.11\tiny{$\pm$2.72 }
&83.81\tiny{$\pm$0.31 }
&74.80\tiny{$\pm$0.32 }
&90.37\tiny{$\pm$0.55 }
&65.91\tiny{$\pm$0.93 }
\\
GRADE &85.67\tiny{$\pm$0.92}
&74.21\tiny{$\pm$0.88}
&83.90\tiny{$\pm$1.27}

&87.17\tiny{$\pm$0.64}
&93.49\tiny{$\pm$0.60}
&82.91\tiny{$\pm$0.83}
\\

\midrule

GT 
&84.32\tiny{$\pm$1.01}
&72.51\tiny{$\pm$1.65}
&\underline{\text{87.77}\tiny{$\pm$}{\text{0.60}}}

&\underline{\text{90.53\tiny{$\pm$2.53}}}
&\underline{\text{95.18}\tiny{$\pm$}{\text{0.66}}}
&{84.05\tiny{$\pm$0.33}}
\\

SAN 
&83.65\tiny{$\pm$1.32}
&72.12\tiny{$\pm$1.89}
&81.04\tiny{$\pm$0.99}

&{90.30\tiny{$\pm$1.06}}
&{95.08\tiny{$\pm$0.48}}
&81.04\tiny{$\pm$0.81}
\\
SAT 
&79.13\tiny{$\pm$0.73}
&66.52\tiny{$\pm$0.60}
&\textbf{\text{87.92}\tiny{$\pm$}\textbf{\text{0.22}}}

&87.78\tiny{$\pm$0.59}  
&92.74\tiny{$\pm$0.51}
&80.04\tiny{$\pm$0.76}
\\

ANS\_GT 
&{86.85\tiny{$\pm$1.34}}
&{74.87\tiny{$\pm$1.53}}
&84.92\tiny{$\pm$0.79}
&88.51\tiny{$\pm$1.74}
&94.29\tiny{$\pm$1.24}
&81.19\tiny{$\pm$2.62}
\\

NodeFormer      
&86.74\tiny{$\pm$0.99}
&72.20\tiny{$\pm$0.79}
&86.24\tiny{$\pm$1.07}
&87.07\tiny{$\pm$1.53}
&93.21\tiny{$\pm$2.71}
&65.47\tiny{$\pm$2.36}
\\ 

UGT  
&\textbf{\text{87.78}\tiny{$\pm$}\textbf{\text{1.85}}}
&\underline{\text{76.24}\tiny{$\pm$}{\text{0.92}}}
&82.98\tiny{$\pm$1.84}

&{{90.18}\tiny{$\pm$}{{1.42}}}
&88.25\tiny{$\pm$1.28}
&\underline{\text{84.12}\tiny{$\pm$}{\text{0.58}}}
\\



\midrule
\text{DegFairGT} 
&\underline{\text{87.10\tiny{$\pm$1.53}}}
&\textbf{\text{76.59}\tiny{$\pm$}\textbf{\text{0.98}}}
&{86.86\tiny$\pm$0.12}

&\textbf{\text{91.45\tiny{$\pm$0.58}}}
&\textbf{\text{95.73}\tiny{$\pm$}\textbf{\text{0.84}}}
&\textbf{\text{84.61\tiny{$\pm$0.53}}}
\\\bottomrule
\end{tabular}
\label{tab:node_classification}
\end{table*}

\subsubsection{Fairness according to Neighborhood Ranges}

The lack of messages of low-degree nodes becomes more severe if their neighbors also do not have enough neighbors.
To evaluate the model fairness under the more severe imbalance, we performed experiments for the imbalance of two-hop neighborhoods ($r = 2$), as shown in Table~\ref{tab:fainess_2_20}. 
(\textbf{i}) DegFairGT outperformed the baselines in terms of both degree fairness metrics in most cases of $r=2$, as with $r=1$.
DegFairGT performed significantly better than the baselines in Cora and Citeseer, which are sparse graphs with the density of $14.81 \times 10^{-4}$ and $8.55 \times 10^{-4}$, respectively.
Also, DegFairGT exhibited better $\Delta_{EO}$ in $r=2$ than $1$.
Specifically, in $r=1$, DegFairGT had smaller $\Delta_{SP}$ than $\Delta_{EO}$, but they were similar in $r=2$. 
Although the more severe imbalance worsened both biases in model prediction and accuracy, the accuracy biases of DegFairGT caught up with the prediction biases in every dataset while still $\Delta_{SP} \leq \Delta_{EO}$ in the baselines. 
This result underpins that community structures are useful for discovering correlated nodes under the severe lack of messages. 
(\textbf{ii}) DegFairGT and other graph transformers, e.g., GT, SAN, SAT, and UGT, exhibited better performance in terms of $\Delta_{EO}$ in $r=2$ than $1$, mostly worse regarding $\Delta_{SP}$. 
However, their performance improvements were focused on Computers, Photo, and WikiCS with the density of $26.00\times10^{-4}$, $40.70\times 10^{-4}$, and $31.57\times 10^{-4}$, respectively, which are more dense than the others.
This implies that in non-sparse graphs when enough neighbors are reachable in $k$-hops, node proximity is effective for detecting correlated nodes.
We can also assume that selecting neighbors satisfying both node proximity and structural similarity makes DegFairGT effectively alleviate degree bias on both sparse and non-sparse graphs. 



\subsubsection{Fairness Analysis on $r=1$ and Top/Bottom 30\%} 
\label{app:C1}

We further evaluate degree fairness by extending $G_1$ and $G_2$ to the top and bottom 30\% nodes, as shown in Table \ref{tab:fainess_1_30}.
This experiment aims to validate whether DegFairGT is also effective for nodes with less severe shortages or overabundances of messages. 
We have the following observations:
\textbf{(i)} DegFairGT outperformed most baselines, as with the 20\% case, and exhibited better performance in 30\% than 20\% excluding $\Delta_{EO}$ in Cora and $\Delta_{SP}$ in Citeseer. 
The decrement on Cora and Citeseer could come from that extending observation ranges for nodes without a severe shortage of messages can cause the noisy signal problem, considering that DegFairGT significantly outperformed the baselines on Cora and Citeseer in the previous experiments.
This was also observed in other graph transformers, e.g., SAT, Graphormer, and UGT. 
\textbf{(ii)} Augmentation-based methods generally exhibited better performance in 30\% and bigger performance improvements from 20\% to 30\% than graph transformers in Cora, Citeseer, and Pubmed, and it was the opposite in Computers, Photo, and WikiCS. 
Similar to the previous experiment, this result underpins the effectiveness of structural similarity and node proximity in sparse and non-sparse graphs, respectively.

\begin{table*}
\centering
\setlength{\tabcolsep}{2.5 pt} 
\caption{The performance on node clustering in terms of Conductance (C) and Modularity (Q).
}
\begin{tabular}{l cc cc cc cc cc cc}
   \toprule
\multirow{1}{*}{}
        &\multicolumn{2}{c}{Cora}&\multicolumn{2}{c}{Citeseer}&\multicolumn{2}{c}{Pubmed}&\multicolumn{2}{c}{Computers}&\multicolumn{2}{c}{Photo}&\multicolumn{2}{c}{WikiCS}
       \\
       \cmidrule(lr){2-3}\cmidrule(lr){4-5}
       \cmidrule(lr){6-7}\cmidrule(lr){8-9}
       \cmidrule(lr){10-11}\cmidrule(lr){12-13}

         &C $\downarrow $& Q $\uparrow $
         &C $\downarrow $& Q $\uparrow $
         &C $\downarrow $& Q $\uparrow $
         &C $\downarrow $& Q $\uparrow $
         &C $\downarrow $& Q $\uparrow $
         &C $\downarrow $& Q $\uparrow $
        \\\midrule


{GIN   }

&22.73\tiny{$\pm$1.50}&59.52\tiny{$\pm$1.93}
&23.95\tiny{$\pm$3.29}&55.35\tiny{$\pm$3.38}
&13.25\tiny{$\pm$0.84}&49.01\tiny{$\pm$0.95}
&39.01\tiny{$\pm$1.42}&59.57\tiny{$\pm$1.19}
&32.69\tiny{$\pm$5.81}&62.53\tiny{$\pm$6.69}
&37.87\tiny{$\pm$3.92}&60.48\tiny{$\pm$3.90}

\\
{GAT   }
&16.05\tiny{$\pm$0.45}
&\underline{{\text{67.29}}{\text{\tiny{$\pm$0.35}}}}

&21.94\tiny{$\pm$0.83}
&58.17\tiny{$\pm$0.69}

&\underline{10.74\tiny{$\pm$0.45}}
&53.21\tiny{$\pm$0.76}

&20.68\tiny{$\pm$2.40}
&77.41\tiny{$\pm$2.08}

&{15.33\tiny{$\pm$0.42}}
&{81.15\tiny{$\pm$0.17}}

&31.78\tiny{$\pm$1.39}
&66.38\tiny{$\pm$2.06}
\\


\midrule
{RGRL}
&12.73\tiny{$\pm$1.44}
&63.92\tiny{$\pm$4.96}
&5.66\tiny{$\pm$0.37}
&\underline{\text{68.41}\text{\tiny{$\pm$3.20}}}
&12.71\tiny{$\pm$1.68}
&{56.36\tiny{$\pm$5.39}}
&{{12.06\tiny{$\pm$1.28}}}
&\underline{{87.76\tiny{$\pm$1.77}}}
&17.57\tiny{$\pm$3.62}
&76.80\tiny{$\pm$5.38}
&\underline{\text{22.62}\text{\tiny{$\pm$1.42}}}
&\textbf{\text{77.47}}\textbf{\text{\tiny{$\pm$2.41}}}\\

{GRACE}

&22.22\tiny{$\pm$0.06}&47.80\tiny{$\pm$0.07}
&\textbf{\text{5.00}}\textbf{\text{\tiny{$\pm$0.04}}}
&\textbf{\text{71.26}}\textbf{\text{\tiny{$\pm$0.04}}}
&14.13\tiny{$\pm$0.04}&48.45\tiny{$\pm$0.03}
&{13.56\tiny{$\pm$0.50}}
&85.09\tiny{$\pm$0.56}
&\textbf{\text{9.32}}\textbf{\text{\tiny{$\pm$0.00}}}
&{85.37\tiny{$\pm$0.00}}
&51.97\tiny{$\pm$0.33}&48.10\tiny{$\pm$1.15}
\\

{GCA}

&{11.76\tiny{$\pm$2.05}}
&61.26\tiny{$\pm$2.64}
&13.52\tiny{$\pm$1.72}
&57.53\tiny{$\pm$1.50}
&20.80\tiny{$\pm$0.07}
&42.23\tiny{$\pm$0.09}
&13.69\tiny{$\pm$1.47}
&{86.21\tiny{$\pm$2.08}}
&21.73\tiny{$\pm$5.63}
&72.20\tiny{$\pm$7.93}
&38.95\tiny{$\pm$6.05}
&60.59\tiny{$\pm$5.55}
\\

{DegFairGNN}

&20.17\tiny{$\pm$1.04}   
&44.78\tiny{$\pm$1.26}

&25.87\tiny{$\pm$0.81}   
&43.31\tiny{$\pm$1.70}

&17.62\tiny{$\pm$1.05} 
&47.59\tiny{$\pm$0.53}

&\underline{ \text{11.87\tiny{$\pm$0.38}}}
&{ \text{87.23\tiny{$\pm$1.96}}}

&{11.99\tiny{$\pm$1.02}}
&\textbf{\text{86.06\tiny{$\pm$1.02}}}

&30.08\tiny{$\pm$0.90} 
&70.47\tiny{$\pm$2.27}
\\

\midrule
GT
&17.77\tiny{$\pm$0.81}&65.07\tiny{$\pm$0.83}
&23.50\tiny{$\pm$0.91}&56.69\tiny{$\pm$1.01}
&19.16\tiny{$\pm$0.99}&44.90\tiny{$\pm$0.85}
&26.53\tiny{$\pm$6.77}&72.77\tiny{$\pm$7.68}
&17.11\tiny{$\pm$0.27}&78.47\tiny{$\pm$0.17}
&34.19\tiny{$\pm$3.57}&64.62\tiny{$\pm$4.18}

\\
SAN

&22.88\tiny{$\pm$2.63}&60.81\tiny{$\pm$2.45}
&24.48\tiny{$\pm$1.95}&56.01\tiny{$\pm$1.77}
&14.89\tiny{$\pm$1.20}&49.72\tiny{$\pm$1.08}
&30.61\tiny{$\pm$6.09}&67.36\tiny{$\pm$6.00}
&18.05\tiny{$\pm$1.71}&73.86\tiny{$\pm$2.63}
&30.22\tiny{$\pm$1.88}&67.95\tiny{$\pm$2.79}
\\

SAT

&28.25\tiny{$\pm$2.47}&54.09\tiny{$\pm$3.00}
&34.82\tiny{$\pm$2.94}&45.31\tiny{$\pm$6.08}
&21.57\tiny{$\pm$1.97}&43.04\tiny{$\pm$2.21}
&20.64\tiny{$\pm$5.15}&79.07\tiny{$\pm$5.95}
&20.67\tiny{$\pm$7.44}&71.85\tiny{$\pm$8.18}
&32.71\tiny{$\pm$3.84}&66.96\tiny{$\pm$4.01}

\\

 
NodeFormer
&15.04\tiny{$\pm$2.36}
&{66.92\tiny{$\pm$1.23}}

&14.82\tiny{$\pm$2.86}
&65.17\tiny{$\pm$1.15}

&14.94\tiny{$\pm$0.51}
&47.89\tiny{$\pm$2.65}

&18.52\tiny{$\pm$0.63}
&80.60\tiny{$\pm$0.43}

&17.04\tiny{$\pm$0.55}
&78.10\tiny{$\pm$0.16}

&{27.07\tiny{$\pm$1.48 }}
&56.88\tiny{$\pm$0.43}
\\
UGT
&\underline{\text{10.28\tiny{$\pm$0.71}}}
&66.23\tiny{$\pm$0.42}
&\underline{\text{5.22\tiny{$\pm$\text{0.94}}}}
&67.18\tiny{$\pm$0.28}
&19.70\tiny{$\pm$5.67}
&\underline{\text{61.40\tiny{$\pm$1.89}}}
&21.83\tiny{$\pm$3.86}
&77.89\tiny{$\pm$4.25}
&16.21\tiny{$\pm$0.56}
&79.00\tiny{$\pm$0.24}
&{27.83\tiny{$\pm$1.27}}
&{72.53\tiny{$\pm$1.44}}
\\

\midrule
\multirow{1}{*}{\text{DegFairGT}}

&\textbf{\text{9.84}}\textbf{\text{\tiny{$\pm$0.76}}}
&\textbf{\text{{69.28}}\textbf{\text{\tiny{$\pm$0.32}}}}

&{5.40\tiny{$\pm$1.21}}
&{68.19\tiny{$\pm$0.39}}

&\textbf{\text{7.66}}\textbf{\text{\tiny{$\pm$1.52}}}
&\textbf{\text{89.51}}\textbf{\text{\tiny{$\pm$4.19}}}

&\textbf{\text{10.13}}\textbf{\text{\tiny{$\pm$1.30}}}
&\textbf{\text{88.07}}\textbf{\text{\tiny{$\pm$1.32}}}

&\underline{\text{9.71}\text{\tiny{$\pm$0.44}}}
&\underline{\text{85.39}\text{\tiny{$\pm$2.58}}}

&\textbf{\text{21.68}}\textbf{\text{\tiny{$\pm$0.82}}}
&\underline{\text{76.71}\text{\tiny{$\pm$1.31}}}
\\
\bottomrule
\end{tabular}
\label{tab:node_clustering}
\end{table*}

\subsection{Performance Analysis}

\subsubsection{Node Classification}




Table~\ref{tab:node_classification} shows the node classification performance of DegFairGT and the baselines.
We have the following observations:
\textbf{(i)} DegFairGT consistently outperformed graph transformers, e.g., GT and SAN, which employ $k$-hop neighborhood sampling regardless of structural similarity. 
This indicates that our approach to refining context nodes with both perspectives is effective for aggregating informative messages and avoiding noisy ones for overall nodes, not only nodes with low or high degrees.
\textbf{(ii)} DegFairGT also outperformed augmentation-based methods, e.g., RGRL and GRADE, exhibiting the power of our learnable augmentation compared to random or heuristic ones.
This indicates that node proximity is significant in node classification to aggregate features from homophilic nodes.
Random or heuristic augmentation cannot easily reflect node proximity since it can generate arbitrary edges. 




\subsubsection{Trade-off between Fairness and Accuracy}

\begin{figure}[t]
\centering
\includegraphics[width= \linewidth]{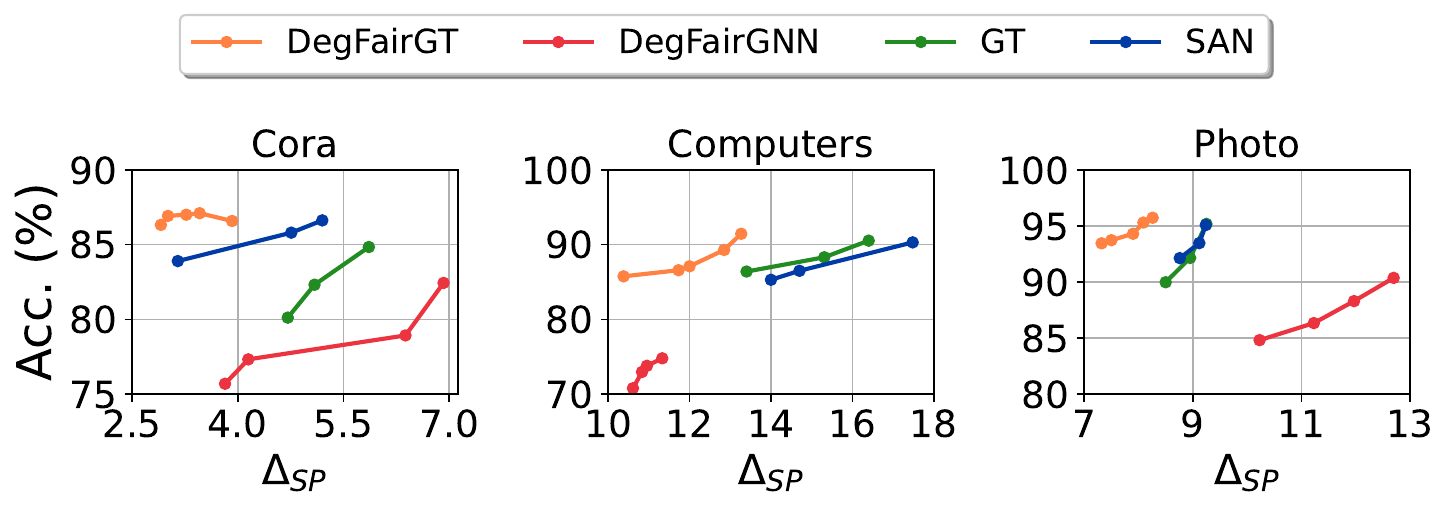}
\caption{The trade-off between Accuracy and $\Delta_{SP}$ in DegFairGT, DegFairGNN, GT, and SAN models on three datasets, e.g., Cora, Computers, and Photo.
The upper left corner holds high accuracy and low $\Delta_{SP}$.}
\label{fig:trade_off}
\end{figure}

We note that there often exists a trade-off between fairness and node classification accuracy \cite{DBLP:conf/icml/BoseH19,DBLP:conf/wsdm/DaiW21}.
Figure \ref{fig:trade_off} presents the Pareto front curves constructed by searching hyper-parameters for each model to compare the trade-off of DegFairGT with the baselines.
The upper-left corner area defines the ideal performance with high accuracy and low degree-related bias.
We observed that the trade-offs between $\Delta_{SP}$ and accuracy are less significant in DegFairGT than in the other models. 
DegFairGNN, which uses heuristic augmentation considering node degrees, showed the most severe trade-off, and GT and SAN, which employ $k$-hop neighborhoods sampling, were between DegFairGNN and DegFairGT. 
The severe trade-off of DegFairGNN underpins that although augmentation considering only structural similarity can alleviate degree bias, we can meet difficulties in discovering homophilic nodes for node classification if we lose node proximity due to uncontrolled edge perturbation. 
The graph transformers show the effectiveness of broader observation ranges based on node proximity, but DegFairGT exhibits that node proximity and structural similarity can be synergetic in discovering correlated nodes. 




\subsubsection{Node Clustering}

Despite the effectiveness of our structural augmentation, the edge perturbation modifies structures of input graphs and could hinder graph structure preservation.
Therefore, we conducted experiments on the node clustering task, as shown in Table \ref{tab:node_clustering}.
\textbf{(i)} DegFairGT outperformed the baselines in most cases, exhibiting the effectiveness of our structure preservation tasks. 
Especially, DegFairGT, UGT, and augmentation-based models, which employ pre-training tasks using contrastive objectives for preserving node connectivity or transition probabilities, showed good performance. 
We assume that DegFairGT outperformed the others since DegFairGT samples nodes satisfying both node proximity and structural similarity in an end-to-end trainable manner.
\textbf{(ii)} Contrary, most GNNs and graph transformers without structure preservation tasks, e.g., GIN, GT, SAN, and SAT, faced difficulties in recovering community structures. 
This implies that only aggregating node features is insufficient to preserve graph structure, especially with a global view.

\subsection{Model Analysis}

\subsubsection{Ablation Analysis}

\begin{table}[tb]
\centering
\setlength{\tabcolsep}{2.5 pt} 
\caption{Ablation study on the presence of the graph augmentation (Aug) and self-attention (Att) modules.
}
\begin{tabular}{cc c ccc}
   \toprule

Aug & Att & Metrics & Cora & Computers &Photo  \\
\midrule

\multirow{4}{*}{\textemdash}
&\multirow{2}{*}{\textemdash} 
&Acc.$\uparrow$
&84.81\tiny{$\pm$1.78}
&{90.07\tiny{$\pm$0.82}}
&94.23\tiny{$\pm$1.45}
\\ 
&&${{\Delta}_{SP}}$$\downarrow$
&5.88\tiny{$\pm$1.09}
&5.96\tiny{$\pm$0.22}
&9.33\tiny{$\pm$0.96}
\\
\cmidrule(lr){2-6}
&\multirow{2}{*}{$\checkmark$} 
&Acc.$\uparrow$
&\underline{{{\text{85.54\tiny{$\pm$1.07}}}}}
&89.01\tiny{$\pm$0.84}
&{95.08\tiny{$\pm$1.08}}\\

&&${{\Delta}_{SP}}$$\downarrow$
&4.85\tiny{$\pm$1.25}
&\underline{\text{{5.45\tiny{$\pm$0.21}}}}
&8.67\tiny{$\pm$0.86}
\\
\cmidrule(lr){1-6}
\multirow{4}{*}{$\checkmark$}
&\multirow{2}{*}{\textemdash} 
&Acc.$\uparrow$
&{84.95\tiny{$\pm$1.19}}
&\underline{{{\text{90.38\tiny{$\pm$0.41}}}}}
&\underline{{{\text{95.46\tiny{$\pm$0.23}}}}}
\\ 
&&${{\Delta}_{SP}}$$\downarrow$
&\underline{\text{{3.53\tiny{$\pm$0.88}}}}
&5.91\tiny{$\pm$0.87}
&\textbf{\text{{8.14\tiny{$\pm$0.68}}}}
\\
\cmidrule(lr){2-6}
&\multirow{3}{*}{$\checkmark$} 
&Acc.$\uparrow$
&\textbf{{\text{87.10\tiny{$\pm$1.53}}}}
&\textbf{{{\text{91.45\tiny{$\pm$0.58}}}}}
&\textbf{{{\text{95.73\tiny{$\pm$0.84}}}}}
\\ 
&&${{\Delta}_{SP}}$$\downarrow$
&\textbf{{\text{3.46\tiny{$\pm$0.97} }}}
&\textbf{\text{{4.99\tiny{$\pm$0.64}}}}
&\underline{\text{{8.26\tiny{$\pm$0.18}}}}
\\
\bottomrule
\end{tabular}
\label{tab:ablation_2}
\end{table}

\begin{table*}[tb]
\centering
\caption{An analysis of the hyper-parameters $\xi$ and $\zeta$.
}
\begin{tabular}{ll ccc ccc ccc }
   \toprule
\multirow{2}{*}{$\xi$}
&\multirow{2}{*}{$\zeta$}

&\multicolumn{3}{c}{Cora} 
& \multicolumn{3}{c}{Computers}
& \multicolumn{3}{c}{Photo}
\\

\cmidrule(lr){3-5}
\cmidrule(lr){6-8}
\cmidrule(lr){9-11}

&
& Acc.$\uparrow$
& ${{\Delta}_{SP}}\downarrow $
&  ${{\Delta}_{EO}}\downarrow $

& Acc.$\uparrow$
& ${{\Delta}_{SP}}\downarrow $
&  ${{\Delta}_{EO}}\downarrow $

& Acc.$\uparrow$
& ${{\Delta}_{SP}}\downarrow $
&  ${{\Delta}_{EO}}\downarrow $
\\
\midrule
\multirow{4}{*}{0.2}
& 0.2
&82.32 
&5.42 
&7.92 
&88.75 
&9.69 
&15.33 
&88.70 
&10.55 
&14.60 
\\
& 0.4
&85.61 
&4.83
&8.01 
&89.33 
&9.06 
&15.37 
&90.47 
&10.12 
&15.48 
\\
& 0.6
&84.84 
&5.78
&9.39 
&85.81 
&9.74 
&16.26 
&88.65 
&10.64 
&15.02 
\\
&0.8
&83.89
&4.48 
&11.76
&84.01
&8.19
&16.63
&87.51
&11.41
&18.95
\\
\midrule
\multirow{4}{*}{0.4}
& 0.2
&82.58 
&5.56 
&6.72 
&84.04 
&8.82 
&13.25 
&88.19 
&9.40 
&18.95 
\\
& 0.4
&84.99
&4.03 
&7.98
&84.65 
&10.79 
&15.88 
&81.43 
&10.26
&17.41 
\\
& 0.6
&85.95
&5.30 
&11.26 
&86.33 
&9.11 
&14.42 
&87.38  
&11.26 
&16.65 
\\
& 0.8
&84.13
&7.48
&10.07 
&85.96
&8.68
&13.61
&87.25
&12.61
&15.76
\\
\midrule
\multirow{4}{*}{0.6}
& 0.2
&84.09 
&5.80 
&10.43
&82.61 
&5.97 
&14.61 
&86.66 
&10.37 
&10.95
\\
& 0.4
&85.02 
&4.07 
&8.25 
&84.64 
&5.66 
&13.69 
&87.90 
&9.61 
&11.28
\\
& 0.6
&86.01
&4.94 
&8.60 
&87.90 
&6.92 
&17.41 
&93.59 
&9.67 
&12.62 
\\
& 0.8
&85.35
&4.10
&9.94 
&84.31
&7.22
&16.33
&89.67
&10.53
&\underline{10.78}
\\
\midrule
\multirow{4}{*}{0.8}
& 0.2 
&\textbf{87.10}
&\textbf{3.46}
&\textbf{4.26}
&\textbf{91.45}
&\textbf{4.48}
&\underline{13.04}
&\textbf{95.73}
&\underline{8.62}
&11.13
\\
& 0.4
&\underline{86.54}
&4.07
&5.10 
&\underline{90.30}
&\underline{4.65}
&\textbf{12.36} 
&\underline{95.69}
&9.12 
&\textbf{10.65}
\\
&0.6
&83.54 
&\underline{3.87}
&\underline{4.56}  
&88.53 
&5.11 
&13.17 
&94.78 
&\textbf{8.54} 
&12.66 
\\
&0.8
&84.17
&4.39
&6.89
&86.21
&5.78
&13.16
&94.53
&9.05
&13.52
\\

\bottomrule
\end{tabular}
\label{tab:balance_xi_etta}
\end{table*}

\begin{table}[tb]
\centering
\setlength{\tabcolsep}{2.5 pt} 
\caption{Sensitivity analysis on the choice of clustering algorithm in terms of Accuracy (Acc.), $\Delta_{SP}$, and $\Delta_{EO}$.
}
\begin{tabular}{l c ccc}
\toprule

Methods 
&Metrics
&Cora 
&Computers
&Photo
\\
\midrule
\multirow{3}{*}{w/o clustering}
&Acc.$\uparrow$
&84.39$\pm$2.14	
&86.95$\pm$1.34
&92.81$\pm$1.21

\\
&${{\Delta}_{SP}}$$\downarrow$
&6.21$\pm$0.82
&6.12$\pm$0.29
&12.41$\pm$1.05
\\
&${{\Delta}_{EO}}$$\downarrow$
&7.34$\pm$1.22	
&16.70$\pm$0.86	
&18.47$\pm$0.95
\\ 

\midrule
\multirow{3}{*}{\makecell[l]{Normalized\\spectral clustering }}
&Acc.$\uparrow$
&\underline{86.63$\pm$1.51}
&\textbf{91.45$\pm$0.59}
&\underline{94.14$\pm$0.64}
\\
&${{\Delta}_{SP}}$$\downarrow$
&\underline{4.48$\pm$0.66}
&\textbf{4.97$\pm$0.56}
&\underline{9.25$\pm$0.91}
\\
&${{\Delta}_{EO}}$$\downarrow$
&\underline{4.98$\pm$1.17}
&\underline{14.55$\pm$1.24}	
&\underline{14.53$\pm$0.73}
\\

\midrule
\multirow{3}{*}{K-mean clustering} 
&Acc.$\uparrow$
&\textbf{87.10$\pm$1.53 }
&\underline{90.38$\pm$1.03}
&\textbf{95.73$\pm$0.84}
\\
&${{\Delta}_{SP}}$$\downarrow$
&\textbf{3.46$\pm$0.97}
&\underline{4.99$\pm$0.64}
&\textbf{8.26$\pm$0.18}
\\
&${{\Delta}_{EO}}$$\downarrow$
&\textbf{4.26$\pm$0.24}	
&\textbf{12.15$\pm$0.84}	
&\textbf{11.13$\pm$3.05}
\\
\bottomrule
\end{tabular}
\label{tab:sen_cluster}
\end{table}

We conducted experiments to validate how much our structural augmentation and structural self-attention modules contribute to overall model performance, as shown in Table~\ref{tab:ablation_2}.
\textbf{(i)} The cases only with the structural augmentation and only with the structural self-attention exhibited similar performance on Cora and Computers.
This result verifies that sampling nodes in the same community with similar roles by using our augmentation module can provide informative messages. 
\textbf{(ii)} DegFairGT with both modules exhibited the best performance in most datasets and metrics. 
This implies that the augmentation and self-attention modules that have the same objectives can complement each other.

\subsubsection{Sensitivity Analysis}

\begin{figure}[tb]
\centering

\begin{subfigure}{\linewidth}
\includegraphics[width=\linewidth]{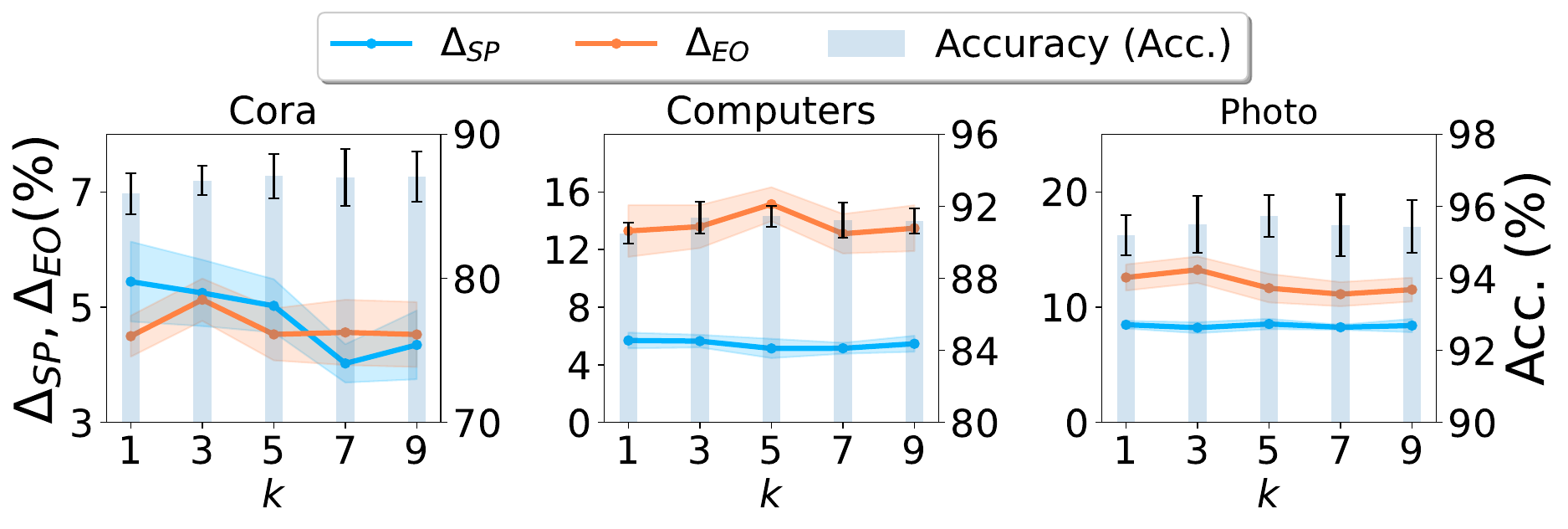}
\caption{On the numbers of clusters ($M$).}
\end{subfigure}
\vskip\baselineskip
\begin{subfigure}{\linewidth}
\includegraphics[width=\linewidth]{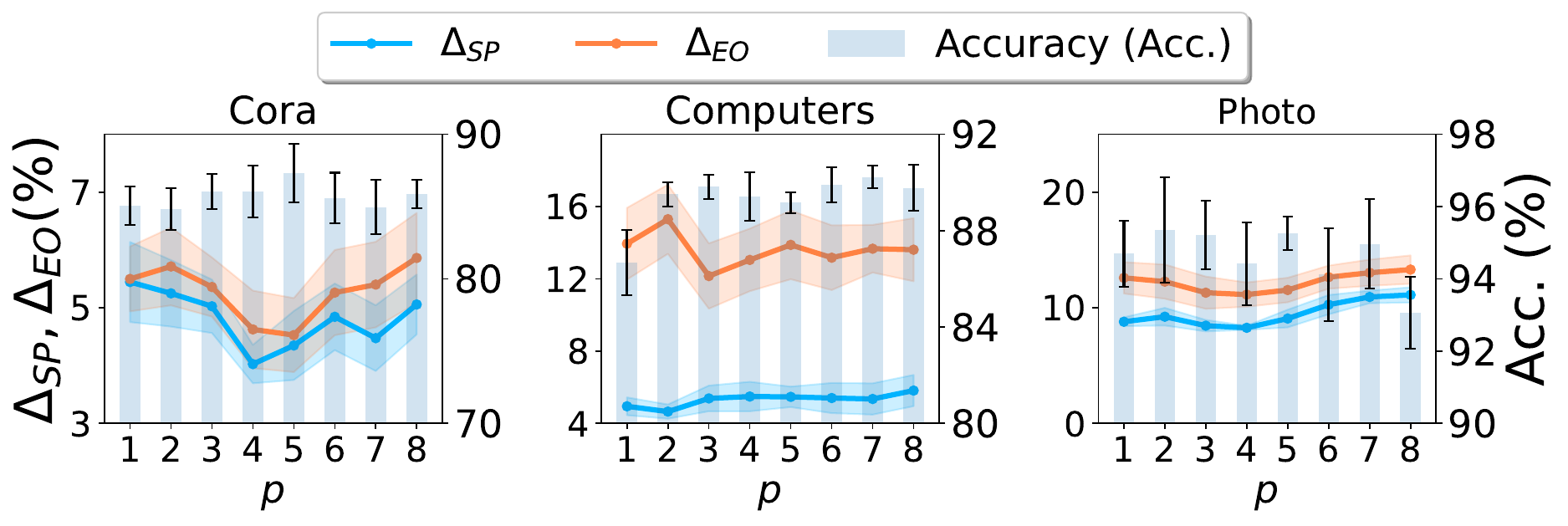}
\caption{On the range of  $p$-step transition.}
\end{subfigure} 
\vskip\baselineskip
\begin{subfigure}{\linewidth}
\includegraphics[width=\linewidth]{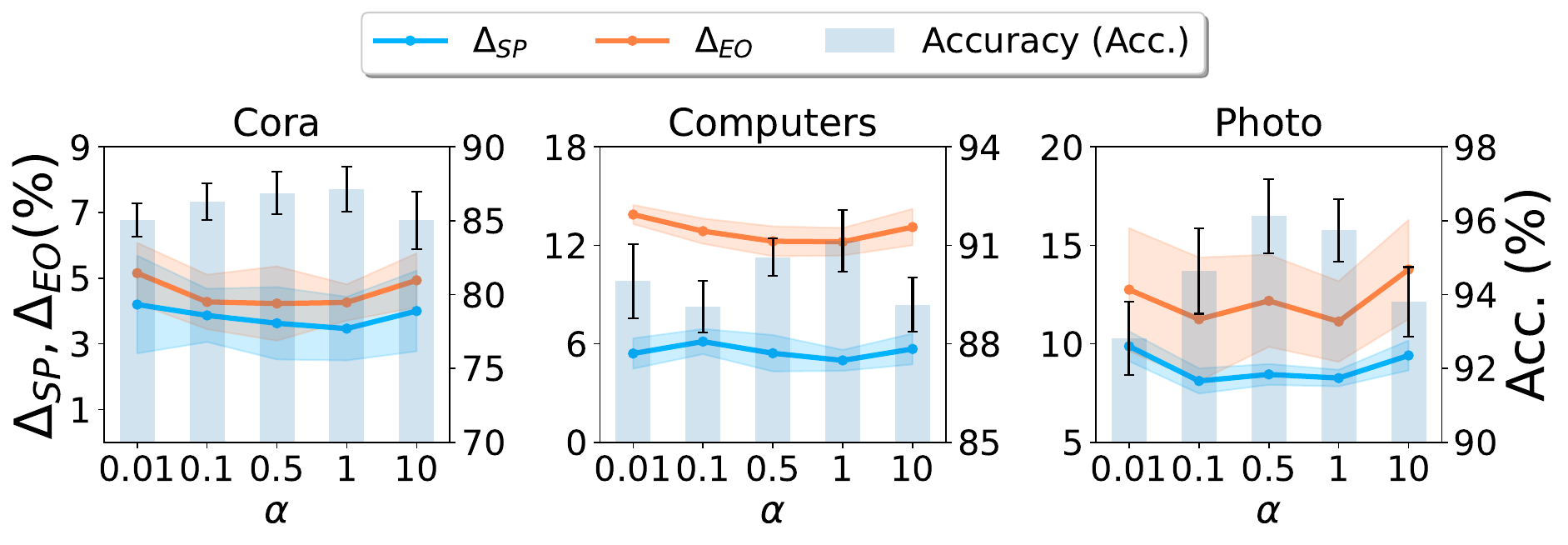}
\caption{On the range of  $\alpha$.}
\end{subfigure}

\caption{Sensitivity analysis on 
(a) the number of clusters,
(b) the range of $p$-step transition preservation, and 
(c) the range of $\alpha$ 
in terms of ${{\Delta}_{SP}}$, ${{\Delta}_{EO}}$, and accuracy on three datasets, e.g., Cora, Computers, and Photo.
}
\label{fig:sas}
\end{figure}

We further investigated the impact of edge sampling on three datasets, i.e., Cora, Computers, and Photo, in terms of accuracy, $\Delta_{SP}$, and $\Delta_{EO}$, as shown in Table \ref{tab:balance_xi_etta}.
The hyperparameters $\xi$ and $\zeta$ are determined with a grid search among $\{ 0.2, 0.4, 0.6, 0.8 \}$.
We observed that DegFairGT has a better performance (higher accuracy and lower $\Delta_{SP}$ and $\Delta_{EO}$) with the $\zeta = 0.2$ and $\xi = 0.8$.
That is, if $\zeta = 0.2$ is large, our structural augmentation can generate noisy edges that could not preserve the original graph structures.
We argue that the structural augmentation could not only generate informative edges but also retain the original graph structures.

As the model performance depends on the choice of clustering algorithms, we investigated how the choice of clustering algorithms affects the performance of DegFairGT in terms of (Acc.), $\Delta_{SP}$, and $\Delta_{EO}$, as shown in Table \ref{tab:sen_cluster}.
We note that spectral clustering captures community structures based on connectivity, ensuring that structurally similar nodes are grouped together, while K-means clustering groups nodes based on feature similarity.
We observed that while normalized spectral clustering can enhance accuracy and fairness effectively, our model with K-means clustering achieved a better balance on three metrics. 
This is because feature similarity is crucial for capturing relationships among non-adjacent nodes, particularly in the context of nodes with similar degrees. We argue that degree-based fairness adjustments in DegFairGT can benefit from clustering methods that prioritize node feature similarities rather than graph connectivity, which aligns with the homophily principle, and similar nodes tend to share common features, even if they are not directly connected. 
By clustering nodes based on feature similarity, K-means better captures meaningful relationships between non-adjacent nodes in the same community.

For the choice of the number of clusters, Figure \ref{fig:sas} (a) illustrates the sensitivity analysis on the number of clusters $M$.
We observed that the model performance remains relatively stable over the $M \ge 3$.
This implies that the model can capture the structural similarity well in the global view to generate informative edges.
Therefore, we select $M=5$ as the number of clusters in all the experiments.




Figure~\ref{fig:sas} (b) presents the performance of DegFairGT according to hyper-parameters $p$-step transition.
\textbf{(i)} $p$ controls the task for $p$-step transition probability preservation, and DegFairGT achieved the best fairness at $p=4$ to $5$ on Cora and at $p=3$ to $4$ on Computers and Photo.
This implies that larger $p$ values allow the model to incorporate global neighborhood information, leading to better fairness representations.
Thus, we can assume that DegFairGT requires a higher $p$ on a more sparse graph to capture global graph structures. 
However, when $p$ is large enough, the global information expands too broadly, which can hurt the model in capturing the local structures.

To validate the contribution of the hyperparameter $\alpha$ in  Equation \ref{eq:total_loss}, we further tuned the hyperparameter $\alpha$, as shown in Figure \ref{fig:sas} (c).
Specifically, we tuned the hyperparameter $\alpha$ among $\{ 0.01, 0.1, 0.5, 1, 10 \}$, following the work \cite{DBLP:conf/cikm/LeeH0P22}.
We observed that our proposed model performed well when $\alpha = 1.0$, the same as the parameter for $L_{1}$.
This implies that both loss functions are equally weighted in guiding the model’s learning process to generate fairness representations.
That is, the model does not excessively change the input graph under our augmentation while preserving the global graph structure with the reconstruction loss.

\begin{table}[tb]
\centering
\setlength{\tabcolsep}{2.5 pt} 
\caption{The computational complexity with Big-O notation.
}
  \begin{tabular}{lc}
    \toprule
     Model & Computational Complexity\\\midrule
     RGRL  &  $O_{projection}(N) $+  $O_{encoder}(KN) $\\
     GRACE  &  $O_{projection }(N) $+  $O_{encoder}(N^2) $\\ 
     \midrule
     GT  &  $O_{PE}(mE) $+  $O_{encoder}(N^2) $\\
     SAT  &   $O_{PE}(mE) $+  $O_{encoder_1}(N^k) $+  $O_{encoder_2}(N^2) $\\
    \midrule
     {DegFairGT} &   $O_{pre}(NE) $+  $O_{bias}(N^2) $ +  $O_{encoder}(N^2) $\\\bottomrule
\end{tabular}
\label{tab:complexity}
\end{table}

\subsubsection{Complexity Analysis}
\label{app:Complexity_Analysis}

Table~\ref{tab:complexity} presents the computational costs of DegFairGT and several baselines.
$O_{projection}(\cdot)$, $O_{PE}(\cdot)$, and $O_{encoder}(\cdot)$ refer to the cost of the projection, positional encoding, and graph encoder.
$m$, $K$, and $E$ present the positional encoding dimension, the number of positive samples for contrastive learning, and the number of edges.
$O_{pre}(\cdot)$ and $O_{bias}(\cdot)$ are the cost of extracting $k$-hop neighborhoods and pre-computing $D$ and $s$ matrices.
The cost of transformer layers of DegFairGT is $O(N^2)$, similar to other graph transformers, such as GT~\cite{dwivedi2020generalization}, while the cost of SAT~\cite{DBLP:conf/icml/ChenOB22} is $O(N^k)$ to aggregate $k$-subgraphs rooted in each node. 
GRACE~\cite{Zhu:2020vf} and RGRL~\cite{DBLP:conf/cikm/LeeH0P22} use GNN encoders with lower computational costs than graph transformers.
In summary, the cost of DegFairGT is lower than SAT and slightly higher than GT due to injecting $s$ into the self-attention with $O(N^2)$.

We note that regarding scalability, the computational complexity of DegFairGT is $O_{encoder}(N^2)$ due to the self-attention mechanism used in the graph transformer encoder.
This means that as the number of nodes in the graph increases, the computational cost rises quadratically, making it infeasible for extremely large graphs.
To address this issue, several techniques, e.g., graph partition or sparse attention \cite{DBLP:conf/iclr/QinSDLWLYKZ22}, could be explored to improve efficiency without compromising performance, which we leave as future work.

\section{Conclusion and Future Works }

This paper proposed a novel degree fairness Graph Transformer, DegFairGT, to address the degree bias of message-passing GNNs via learnable structural augmentation and self-attention.
Instead of randomly sampling node pairs to make edges, we discovered nodes with high structural similarity, i.e., nodes with similar roles in the same community, likely to be sampled to generate edges under our learnable structural augmentation.
By doing so, the model could not only mitigate the degree bias but also preserve the graph structure under our structural self-attention.
The experiments on various benchmark datasets demonstrate the superiority of our proposed model, showing that our framework could achieve degree fairness, classification accuracy, and structure preservation simultaneously.

There are several limitations in our work. First, the main limitation of DegFairGT is its computational inefficiency, as it inherits the quadratic complexity of self-attention, leading to $O{(N^2)}$ complexity, where $N$ is the number of nodes. 
This quadratic scaling becomes computationally expensive for large graphs, requiring significant memory and processing power. 
Second, we focus solely on the roles of nodes (node degree) and their communities to identify structural similarity between node pairs. 
However, there exist other forms of structural similarity, such as semantic subgraph-level similarity, which also play a crucial role in capturing node pair relationships. 
By discovering the high-order structures, the model could better capture diverse forms of structural similarity beyond node degree and community membership

To address the above limitations, future research could explore the following directions: (1) To address computational complexity, we suggest that future research focus on enhancing self-attention mechanisms' computational complexity.  Several recent studies have leveraged improvements in developing linear transformers, which offer linear scalability in computational complexity \cite{DBLP:journals/corr/abs-2009-06732,DBLP:conf/iclr/QinSDLWLYKZ22}.
The follow-up study will aim to improve the computational efficiency and time complexity of self-attention mechanisms by leveraging linear self-attention techniques. 
(2) There exist other forms of structural similarity, such as semantic subgraph-level similarity, which also play a crucial role in capturing node pair relationships. Discovering semantic subgraph-level relationships would enhance the model’s ability to capture more high-order structural relationships between nodes.  By identifying and leveraging shared substructures, e.g., motifs, the model could better represent complex dependencies that are not captured by degree and community alone.

\appendices

\ifCLASSOPTIONcompsoc
  \section*{Acknowledgments}
\else
  \section*{Acknowledgment}
\fi


This work was supported 
in part by the National Research Foundation of Korea (NRF) grant funded by the Korea government (MSIT) (No. 2022R1F1A1065516 and No. 2022K1A3A1A79089461) (O.-J.L.) 
%
%
%
and
in part by the R\&D project “Development of a Next-Generation Data Assimilation System by the Korea Institute of Atmospheric Prediction System (KIAPS)”, funded by the Korea Meteorological Administration (KMA2020-02211) (H.-J.J.).

\ifCLASSOPTIONcaptionsoff
  \newpage
\fi

\end{document}